\documentclass{article}

  \usepackage[eandd, final]{neurips_2026}
\usepackage[utf8]{inputenc} % allow utf-8 input
\usepackage[T1]{fontenc}    % use 8-bit T1 fonts
\usepackage{hyperref}       % hyperlinks
\usepackage{url}            % simple URL typesetting
\usepackage{booktabs}       % professional-quality tables
\usepackage{amsfonts}       % blackboard math symbols
\usepackage{amsmath}
\usepackage{nicefrac}       % compact symbols for 1/2, etc.
\usepackage{microtype}      % microtypography
\usepackage{xcolor}         % colors
\usepackage{graphicx}
\usepackage{multirow}
\usepackage{natbib}
\usepackage{array}
\usepackage{colortbl}
\usepackage{wrapfig}
\usepackage{amsthm}
\usepackage{todonotes}

\title{\rev{When} Predicting Nothing Beats SAM 3: \\Revisiting Evaluation in Video Object Segmentation}

\newcommand{\jnf}{$\mathcal{J}\&\mathcal{F}$}
\newcommand{\jnfv}{$\mathcal{J}\&\mathcal{F}_{v}$}
\newcommand{\mathjnf}{\mathcal{J}\&\mathcal{F}}

\newcommand{\benchname}{FaVOS}
\newcommand{\benchnamefull}{\textbf{FaVOS} (A Benchmark for \textbf{V}ideo \textbf{O}bject \textbf{S}egmentation with \textbf{F}r\textbf{a}ctional Temporal Visibility)}
\newcommand{\pzero}{true negative rate}
\newcommand{\pone}{true positive rate}
\newcommand{\s}{conditional segmentation quality}

\newif\ifshowrev
\showrevfalse
\ifshowrev
  \newcommand{\revcolor}{\color{blue}}
\else
  \newcommand{\revcolor}{}
\fi
\newcommand{\rev}[1]{{\revcolor#1}}
\newenvironment{revblock}{\revcolor}{\par}
\newif\ifshowcorr
\showcorrfalse
\ifshowcorr
  \newcommand{\corrcolor}{\color{blue}}
\else
  \newcommand{\corrcolor}{}
\fi
\newcommand{\corr}[1]{{\corrcolor#1}}

\author{%
  \mdseries
  Jihwan Hong\textsuperscript{1} \quad
  Woohyeon Park\textsuperscript{2} \quad
  Jaeik Kim\textsuperscript{1} \quad
  Jaeyoung Do\textsuperscript{1,2,\dag} \\
  AIDAS Laboratory, \textsuperscript{1}IPAI \& \textsuperscript{2}ECE, Seoul National University \\
  \texttt{\{csjihwanh, woohyeon, jake630, jaeyoung.do\}@snu.ac.kr}
}

\begin{document}

\maketitle
{\renewcommand{\thefootnote}{\fnsymbol{footnote}}\footnotetext[2]{Corresponding author.}}

\begin{abstract}
Video Object Segmentation (VOS) in complex and long videos is increasingly important for real-world applications, where target objects often appear only intermittently within long temporal horizons. However, existing benchmarks largely focus on temporally salient objects that remain visible for most of the video. To address this gap, we introduce \benchnamefull{}, a benchmark designed to evaluate VOS methods under low temporal visibility. We show that, in this regime, the standard \jnf{} metric can collapse VOS evaluation into absence classification, because empty predictions receive high rewards on target-absent frames. Consequently, even a trivial empty-mask predictor can outperform strong models such as SAM~3, revealing a fundamental mismatch between current metrics and practical VOS performance. To mitigate this issue, we propose Volumetric \jnf{}, which evaluates mask sequences as spatio-temporal volumes and reduces the dominance of target-absence rewards while preserving sensitivity to segmentation quality and temporal structure. \rev{Project page: \url{https://aidaslab.github.io/FaVOS}.}

\end{abstract}

\section{Introduction}

\begin{figure}[h]
  % \vspace{-10pt}
  \centering
  \includegraphics[width=\linewidth]{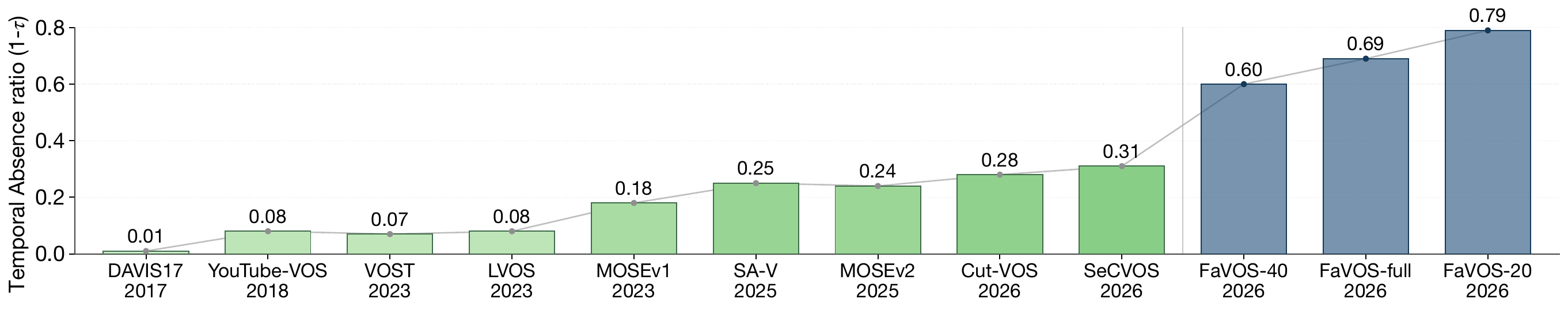}
  % \vspace{-16pt}
  \caption{Temporal absence ratio ($1-\tau$) across VOS benchmarks. Higher values indicate a larger fraction of frames where the target is absent, reflecting lower temporal visibility in recent datasets.}
  % \vspace{-4pt}
  \label{fig:temp_vis}
\end{figure}

Video Object Segmentation (VOS)~\cite{perazzi2016davis,xu2018ytvos} aims to segment target objects in videos at the pixel level. As video understanding systems move toward longer and more realistic inputs, VOS has become increasingly important for crucial domains such as autonomous driving~\cite{yao2020vos_survey,siam2021vcas_driving}, medical imaging~\cite{ji2022videopolyp, zhu2024medsam2}, surveillance~\cite{gao2023vos_survey2,patil2021surveillance}, \rev{wildlife monitoring~\cite{wasmuht2026sa}, egocentric video~\cite{darkhalil2022visor}, and robotic manipulation~\cite{huang2023voxposer}}. In these settings, the target object is not guaranteed to remain continuously visible. It may leave the scene, become fully occluded, reappear after a long interval, or be visible only in a small fraction of the video. A practical VOS system must therefore segment the target accurately when it is visible while avoiding hallucinated masks when it is absent.

Recent VOS benchmarks have increasingly reflected this shift toward long-term and complex videos, including long-term video segmentation~\cite{hong2023lvosv1,hong2025lvosv2} and scenarios involving severe occlusion or disappearance~\cite{mose,mosev2,zhang2025secvos}. To make this factor explicit, we define \textit{temporal visibility}, denoted by $\tau$, which measures the fraction of frames in which the target object is visible. 
As shown in Fig.~\ref{fig:temp_vis}, recent benchmarks tend to contain a larger proportion of target-absent frames, corresponding to lower temporal visibility. This trend suggests that low-visibility VOS is not a corner case, but an increasingly relevant regime for evaluating long-horizon video segmentation systems.

However, we find that standard VOS evaluation is not well calibrated for this regime. The dominant metric, \jnf{}~\cite{perazzi2016davis}, averages region similarity ($\mathcal{J}$) and contour accuracy ($\mathcal{F}$) over frames and is widely used across VOS tasks~\cite{perazzi2016davis,khoreva2018refdavis,gavrilyuk2018a2dsentences,ravi2024sam2,zhou2022avos}. This frame-wise design is natural when the object is usually present, but creates a different scoring behavior when the ground-truth mask is empty: $\mathcal{J}$ and $\mathcal{F}$ no longer measure segmentation quality, but reduce to checking whether the prediction is also empty. Thus, an empty prediction receives a perfect score, whereas even a one-pixel prediction is severely penalized. As temporal visibility decreases, target-absent frames occupy a larger fraction of evaluation, and the average \jnf{} score becomes increasingly determined by absence classification rather than visible-frame segmentation quality, effectively collapsing the task into absence classification.

\begin{figure}[t]

  \centering
  % \vspace{-10pt}
  \includegraphics[width=0.9\linewidth]{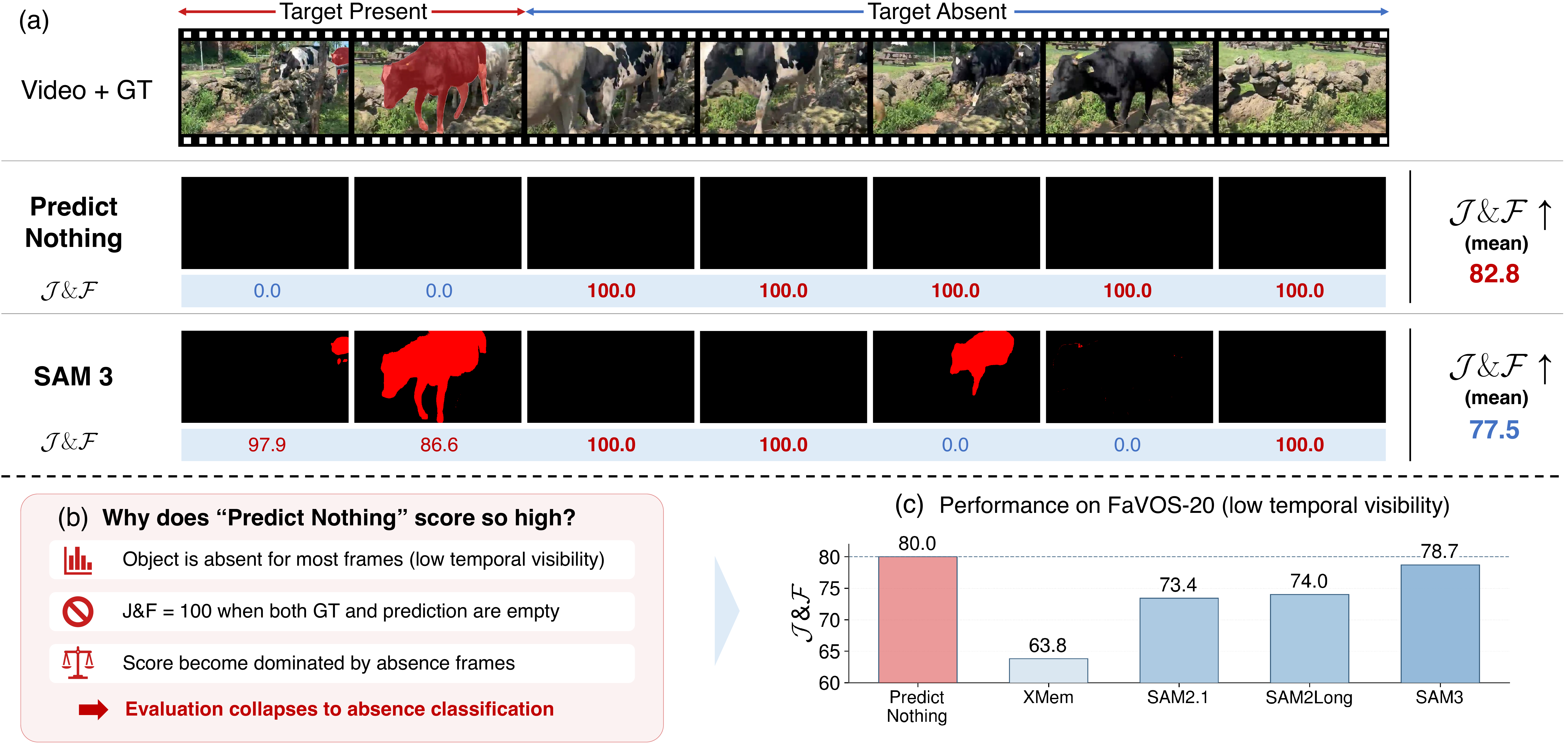}
    % \vspace{-8pt}

    \caption{The standard VOS metric, \jnf{}, assigns perfect scores to empty predictions, allowing a trivial “predict nothing” baseline to outperform strong models under low temporal visibility.}
  \label{fig:main_fig}
  \vspace{-1.0em}
  
\end{figure}

To systematically analyze this issue, we first study the behavior of \jnf{} under varying temporal visibility, and show that its score can become dominated by the absence-classification term. We then introduce a new benchmark, \benchnamefull{}, designed to explicitly control temporal visibility. \benchname{} consists of 200 videos with \corr{279} objects and is divided into two subsets, FaVOS-20 and FaVOS-40, where the object temporal visibility is controlled at 0.2 and 0.4, respectively. Our empirical results reveal a striking consequence of this evaluation mismatch. Under low temporal visibility, \jnf{} strongly favors models with high target-absence accuracy, even when their visible-frame segmentation quality is lower. In the extreme case, a trivial empty-mask predictor, which never segments any object, can match or outperform strong VOS models under \jnf{} on low-visibility subsets. This phenomenon, summarized in Fig.~~\ref{fig:main_fig}, does not imply that empty prediction is a useful VOS strategy; rather, it exposes that the evaluation metric can partially collapse into an absence-classification benchmark.

Finally, we investigate existing metrics designed for temporal consistency, including $\mathcal{T}$~\cite{perazzi2016davis}, $\text{tIoU}$, $\text{vIoU}$~\cite{liang2026longrvos}, and $\mathcal{J}\&\dot{\mathcal{F}}_d$, $\mathcal{J}\&\dot{\mathcal{F}}_r$~\cite{mosev2}. While these metrics capture aspects of temporal stability, temporal presence overlap, or disappearance--reappearance behavior, they do not fully address the collapse of \jnf{} into absence classification while preserving sensitivity to segmentation quality. We therefore propose \textbf{volumetric \jnf{}} (\jnfv{}), which evaluates predicted and ground-truth mask sequences as spatio-temporal volumes. By aggregating overlap over the video volume rather than assigning perfect frame-wise scores to empty frames, \jnfv{} reduces the dominance of absence classification while retaining sensitivity to segmentation quality and temporal consistency.

Our contributions are as follows:
\begin{itemize}

    \item We identify and formalize a low-temporal-visibility failure mode of standard VOS evaluation, showing both theoretically and empirically that frame-wise \jnf{} can become dominated by absence classification rather than visible-frame segmentation quality.

    \item We present \benchnamefull{}, a diagnostic benchmark that explicitly controls temporal visibility, enabling analysis of existing VOS models and evaluation metrics under low-visibility scenarios.
    
    \item We propose \textbf{volumetric \jnf{}} (\jnfv{}), a new spatio-temporal metric that preserves the region and contour formulation of \jnf{} while reducing the dominance of absence classification through volume-level aggregation.
    
\end{itemize}

\section{Problem Formulation}

\subsection{Video Object Segmentation}

Given a video $V = \{v_t\}_{t=1}^{T}$, where $v_t$ denotes the $t$-th frame and $T$ is the number of frames, and first-frame annotations $\{\mathcal{M}_1^{o_i}\}_{i=1}^{N}$ for $N$ objects, video object segmentation aims to propagate object masks over time. Each object is represented as a sequence of binary masks:
\begin{equation}
    \mathcal{M}^{o_i} = \{\mathcal{M}_t^{o_i}\}_{t=1}^{T}, \quad 
    \mathcal{M}_t^{o_i} \in \{0,1\}^{H \times W}.
\end{equation}

Given predicted masks $\hat{\mathcal{M}}^{o_i} = \{\hat{\mathcal{M}}_t^{o_i}\}_{t=1}^{T}$, the objective is to produce masks that are spatially accurate and temporally consistent throughout the video.

\subsection{\jnf}
\label{sec:jnf_definition}

For a ground-truth mask $\mathcal{M}_t^{o_i}$ and prediction $\hat{\mathcal{M}}_t^{o_i}$, let $\|\cdot\|_0$ denote the number of foreground pixels. The Jaccard index is defined as
\begin{equation}
\label{eq:iou_definition}
\mathcal{J}(\mathcal{M}_t^{o_i}, \hat{\mathcal{M}}_t^{o_i})
=
\begin{cases}
\frac{\|\mathcal{M}_t^{o_i} \cap \hat{\mathcal{M}}_t^{o_i}\|_0}
     {\|\mathcal{M}_t^{o_i} \cup \hat{\mathcal{M}}_t^{o_i}\|_0},
& \text{if } \|\mathcal{M}_t^{o_i}\|_0 > 0, \\[4pt]
\mathbf{1}(\|\hat{\mathcal{M}}_t^{o_i}\|_0 = 0),
& \text{otherwise},
\end{cases}
\end{equation}
while contour accuracy $\mathcal{F}(\mathcal{M}_t^{o_i}, \hat{\mathcal{M}}_t^{o_i})$ is defined through boundary precision and recall. When $\|\mathcal{M}_t^{o_i}\|_0 = 0$, $\mathcal{F}$ also reduces to a binary correctness indicator based on whether the prediction is empty. Details of $\mathcal{F}$ are provided in Appendix~\ref{app:f_measure}.

By slight abuse of notation, we use
$\mathcal{J}(\mathcal{M}^{o_i}, \hat{\mathcal{M}}^{o_i})$
and
$\mathcal{F}(\mathcal{M}^{o_i}, \hat{\mathcal{M}}^{o_i})$
to denote temporal averages:
\begin{equation}
\mathcal{J}(\mathcal{M}^{o_i}, \hat{\mathcal{M}}^{o_i})
=
\frac{1}{T}
\sum_{t=1}^{T}
\mathcal{J}(\mathcal{M}_t^{o_i}, \hat{\mathcal{M}}_t^{o_i}),
\qquad
\mathcal{F}(\mathcal{M}^{o_i}, \hat{\mathcal{M}}^{o_i})
=
\frac{1}{T}
\sum_{t=1}^{T}
\mathcal{F}(\mathcal{M}_t^{o_i}, \hat{\mathcal{M}}_t^{o_i}).
\end{equation}

The \jnf{}~\cite{perazzi2016davis} score for object $o_i$ is
\begin{equation}
\mathjnf^{o_i}
=
\frac{1}{2}
\Big(
\mathcal{J}(\mathcal{M}^{o_i}, \hat{\mathcal{M}}^{o_i})
+
\mathcal{F}(\mathcal{M}^{o_i}, \hat{\mathcal{M}}^{o_i})
\Big).
\label{eq4}
\end{equation}

The final \jnf{} score over $K$ videos is computed by averaging over all objects:
\begin{equation}
\mathjnf
=
\frac{1}{\sum_{k=1}^{K} N_k}
\sum_{k=1}^{K}
\sum_{i=1}^{N_k}
\mathjnf^{o_i}.
\label{eq5}
\end{equation}

\subsection{Temporal Visibility}

We quantify whether the target object is present at each frame using a binary visibility indicator. Specifically, the ground-truth visibility at frame $t$ is defined as
\begin{equation}
e_t = \mathbf{1}\left(\|\mathcal{M}_t\|_0 > 0\right),
\label{eq6}
\end{equation}
where $e_t = 1$ if the target object is present, and $0$ otherwise.
The temporal visibility ratio is then defined as
\begin{equation}
\tau(V, \mathcal{M})
=
\frac{1}{T}
\sum_{t=1}^{T} e_t,
\label{eq7}
\end{equation}
i.e., the fraction of frames in which the target object is visible.

\subsection{Analytical Decomposition of VOS Predictors}

To analyze how existing methods and \jnf{} behave under low temporal visibility, we introduce a simplified analytical decomposition of VOS predictors into three components: \pzero{} $p_0$, \pone{} $p_1$, and \s{} $s$. This decomposition is not intended to model every possible VOS error mode or represent a specific architecture. Instead, it serves as a diagnostic abstraction that separates target-absence correctness from visible-frame segmentation quality and analyzes how these components contribute to the final \jnf{} score.

Let $e_t,\hat e_t\in\{0,1\}$ denote the ground-truth and predicted visibility at frame $t$, respectively, where the value $1$ indicates target presence. With $\mathcal{T}_1=\{t:e_t=1\}$ and $\mathcal{T}_0=\{t:e_t=0\}$, we define
\begin{equation}
p_1 =
\frac{1}{|\mathcal{T}_1|}
\sum_{t\in\mathcal{T}_1} \hat e_t,
\qquad
p_0 =
\frac{1}{|\mathcal{T}_0|}
\sum_{t\in\mathcal{T}_0} (1-\hat e_t),
\label{eq8}
\end{equation}
where $p_1$ measures correctness on target-present frames and $p_0$ measures correctness on target-absent frames.

Under this abstraction, the predictor outputs an all-zero mask when $\hat e_t=0$, and outputs a non-empty segmentation mask $\hat{\mathcal{M}}_t$ when $\hat e_t=1$. We further define $s$ as the expected segmentation quality when the object is present and correctly predicted:
\begin{equation}
s :=
\mathbb{E}
\left[
\frac{1}{2}
\left(
\mathcal{J}(\mathcal{M}_t,\hat{\mathcal{M}}_t)
+
\mathcal{F}(\mathcal{M}_t,\hat{\mathcal{M}}_t)
\right)
\,\middle|\,
e_t=1,\hat e_t=1
\right].
\label{eq9}
\end{equation}

Then, the expected \jnf{} score for an object sequence decomposes as
\begin{equation}
\label{eq:jnf_decomposition}
\mathjnf
=
\tau p_1 s + (1-\tau)p_0,
\end{equation}
where $\tau$ denotes the average temporal visibility of an object. The first term corresponds to segmentation quality on target-present frames, while the second term captures absence classification on target-absent frames. The derivation is provided in Appendix~\ref{app:derivation_jnf_decomposition}.

\paragraph{Empty-mask predictor.}
We define the \emph{empty-mask predictor} as a predictor that always outputs the all-zero mask for every frame. Since it never predicts the target object as present, it has $p_0=1$, $p_1=0$, and $s=0$. Substituting these values into Eq.~\ref{eq:jnf_decomposition} yields
\begin{equation}
\mathjnf = 1-\tau.
\end{equation}

Thus, as temporal visibility decreases, the empty-mask predictor can achieve a high \jnf{} score purely by matching target-absent frames, without performing meaningful segmentation.

\section{Related Works}

\subsection{Benchmarks on Video Object Segmentation (VOS)}

Early VOS benchmarks, such as DAVIS~\cite{perazzi2016davis}, established the standard evaluation protocol for pixel-level video segmentation. This was followed by larger-scale datasets such as YouTube-VOS~\cite{xu2018ytvos}, which significantly expanded data diversity and scale. 
More recent benchmarks have focused on increasingly challenging and realistic scenarios. These include distinguishing targets from distractors~\cite{videnovic2025didi}, long-term videos~\cite{hong2023lvosv1,hong2025lvosv2}, handling object transformations~\cite{tokmakov2023vost,chen2025m3vos}, severe occlusions and disappearance~[\citenum{mose}, \citenum{mosev2}, \citenum{zhang2025secvos}, \rev{\citenum{miao2024ovos}}], and multi-cut video structures~\cite{wei2022youmvos,hu2026cutvos}. 

\begin{table}[t]
  \centering
  \setlength{\tabcolsep}{9pt}
  \renewcommand{\arraystretch}{0.95}
  \scriptsize
    \caption{Comparison of video benchmarks in terms of scale and temporal dynamics. Metrics are computed on evaluation splits (test if available; otherwise validation). Disappearance rate~\cite{mose} denotes the fraction of objects absent in at least one frame. * indicates values estimated with fps=6.}
  
  \begin{tabular}{
    l c c c c c 
    >{\centering\arraybackslash\columncolor{blue!8}}p{2.2cm}
  }
    \toprule
    \textbf{Benchmark} &
    \textbf{Year} &
    \textbf{Videos} &
    \textbf{Objects} &
    \textbf{Duration (s)} &
    \textbf{Disappearance Rate} &
    \textbf{Avg. Temporal Visibility} \\
    \midrule

    \multicolumn{7}{c}{\textit{Tracking benchmarks (VOT)}} \\
    \midrule
    GOT-10k~\cite{huang2019got10k} & 2019 & 180 & 180 & 12.7 & 0.00\% & 1.00 \\
    LaSOT~\cite{fan2019lasot} & 2019 & 280 & 280 & 81.6 & 58.6\% & 0.97 \\
    VOT2022~\cite{kristan2022vot2022} & 2022 & 62 & 62 & 10.7 & 19.4\% & 1.00 \\
    DiDi~\cite{videnovic2025didi} & 2025 & 180 & 180 & 254.5* & 42.8\% & 0.97 \\

    \midrule
    \multicolumn{7}{c}{\textit{Segmentation benchmarks (VOS)}} \\
    \midrule
    DAVIS17~\cite{perazzi2016davis} & 2017 & 30 & 61 & 2.8 & 6.6\% & 0.99 \\
    YouTube-VOS~\cite{xu2018ytvos} & 2018 & 474 & 894 & 4.4 & 18.1\% & 0.92 \\
    VOST~\cite{tokmakov2023vost} & 2023 & 70 & 185 & 22.3 & 55.7\% & 0.93 \\
    LVOS~\cite{hong2023lvosv1} & 2023 & 140 & 239 & 78.6 & 58.6\% & 0.92 \\
    MOSE~\cite{mose}   & 2023 & 311 & 630 & 8.7$^*$ & 47.8\% & 0.82 \\
    SA-V~\cite{ravi2024sam2}  & 2025 & 150 & 278 & 17.8 & 84.9\% & 0.75 \\
    MOSEv2~\cite{mosev2} & 2025 & 433 & 575 & 25.6$^*$ & 84.2\% & 0.76 \\
    Cut-VOS~\cite{hu2026cutvos} & 2026 & 100 & 173 & 15.9 & 87.3\% & 0.72 \\
    SeCVOS~\cite{zhang2025secvos}  & 2026 & 160 & 182 & 29.4 & 94.0\% & 0.69 \\
    \midrule
    Ours (FaVOS-full) & 2026 & 200 & \corr{279} & 39.9 & 99.3\% & \textbf{0.31} \\
    Ours (FaVOS-40) & 2026  & 100 & \corr{150} & 36.8 & \corr{98.7\%} & \textbf{\corr{0.40}} \\
    Ours (FaVOS-20) & 2026  & 100 & 129 & 43.0 & 100.0\% & \textbf{0.21} \\
    \bottomrule
  \end{tabular}
  \vspace{-2.0em}
  \label{tab:benchmark-comparison}
\end{table}

Despite these advances, existing benchmarks predominantly focus on \textit{temporally salient} objects, where targets remain visible for a large portion of the video. As shown in Tab.~\ref{tab:benchmark-comparison}, the average temporal visibility across benchmarks remains high, although recent datasets exhibit a decreasing trend. The detailed distribution of temporal visibility is provided in Appendix~\ref{app:benchmark_detail}.

\subsection{Evaluation Metrics for Video Object Segmentation}
\label{subsec3.2}

The standard evaluation metric for VOS is \jnf{}, which combines region similarity $\mathcal{J}$ based on intersection-over-union (IoU)~\cite{everingham2010pascalvoc} and contour accuracy $\mathcal{F}$ following the DAVIS protocol~\cite{perazzi2016davis}. It is widely adopted across semi-supervised~\cite{caelles2017osvos,perazzi2016davis}, unsupervised~\cite{caelles2019uvos}, referring~\cite{khoreva2018refdavis,gavrilyuk2018a2dsentences,ding2023mevis}, and audio-guided VOS~\cite{zhou2022avos,pan2022wnet,ding2025mevisv2}. However, because \jnf{} is computed independently on each frame and averaged over time, it cannot fully capture temporal structure such as disappearance, reappearance, and prediction stability.

Several alternative metrics have been proposed to address temporal behavior. Temporal stability $\mathcal{T}$~\cite{perazzi2016davis} measures mask smoothness across consecutive frames. MOSEv2~\cite{mosev2} introduces $\mathcal{J}\&\dot{\mathcal{F}}$, $\mathcal{J}\&\dot{\mathcal{F}}_d$, and $\mathcal{J}\&\dot{\mathcal{F}}_r$ for small objects and disappearance--reappearance dynamics, but these metrics do not explicitly account for interval duration. Long-RVOS~\cite{liang2026longrvos} further introduces tIoU and vIoU to measure temporal overlap and volumetric consistency. Related volumetric measures are also used in video instance segmentation (VIS)~\cite{yang2019vis}, typically based on volumetric IoU and average precision. However, both vIoU-style and VIS-style metrics do not preserve the contour-based formulation of \jnf{}. Detailed definitions of alternative metrics are provided in Appendix~\ref{app:alternative_jnf}.
\vspace{-0.3em}
\section{FaVOS}
\subsection{Benchmark Construction}

When constructing \benchnamefull{}, we follow three design principles: 
(1) the benchmark construction and evaluation protocol should be reproducible, excluding sources with restrictive licenses or unstable availability (e.g., YouTube videos);
(2) the data distribution should remain close to existing VOS benchmarks; 
and (3) the benchmark should cover diverse real-world scenarios.

Accordingly, we construct the benchmark from two complementary sources: (i) the YFCC100M~\cite{thomee2016yfcc100m} collection and (ii) additional videos collected to increase scenario diversity.
From YFCC100M, we sample 160K videos and apply automatic filtering by removing videos shorter than 20 seconds, low-motion sequences identified via optical flow, and single-object-dominant videos, resulting in approximately 5K candidates. We then manually curate and annotate a subset, excluding low-quality content such as concatenated images, multi-clip edits, fast-rolling footage, infographics, animations, and trivial cases where the target is easily segmentable from the first frame alone. This process yields 133 videos.
We further collect and process additional videos using the same filtering and annotation pipeline, yielding another 67 videos. All annotations are manually verified, and human faces are automatically blurred to address privacy concerns.

\subsection{Temporal Visibility-Controlled Design}

We select target objects that are clearly distinguishable and non-trivial, prioritizing cases with intermittent visibility, including disappearance and reappearance.
After annotation, we construct two subsets by controlling temporal visibility, defined as the fraction of frames in which the target object is visible. Specifically, we build \benchname{}-20 and \benchname{}-40, where temporal visibility is set to 0.2 and 0.4, respectively. This enables controlled analysis of evaluation metrics under low-visibility conditions while maintaining realistic data distribution. Examples are shown in Fig.~\ref{fig:favos_example}; additional examples and benchmark construction details are provided in Appendix~\ref{app:benchmark_detail}.

\begin{figure}[t]
  \centering
  \includegraphics[width=\linewidth]{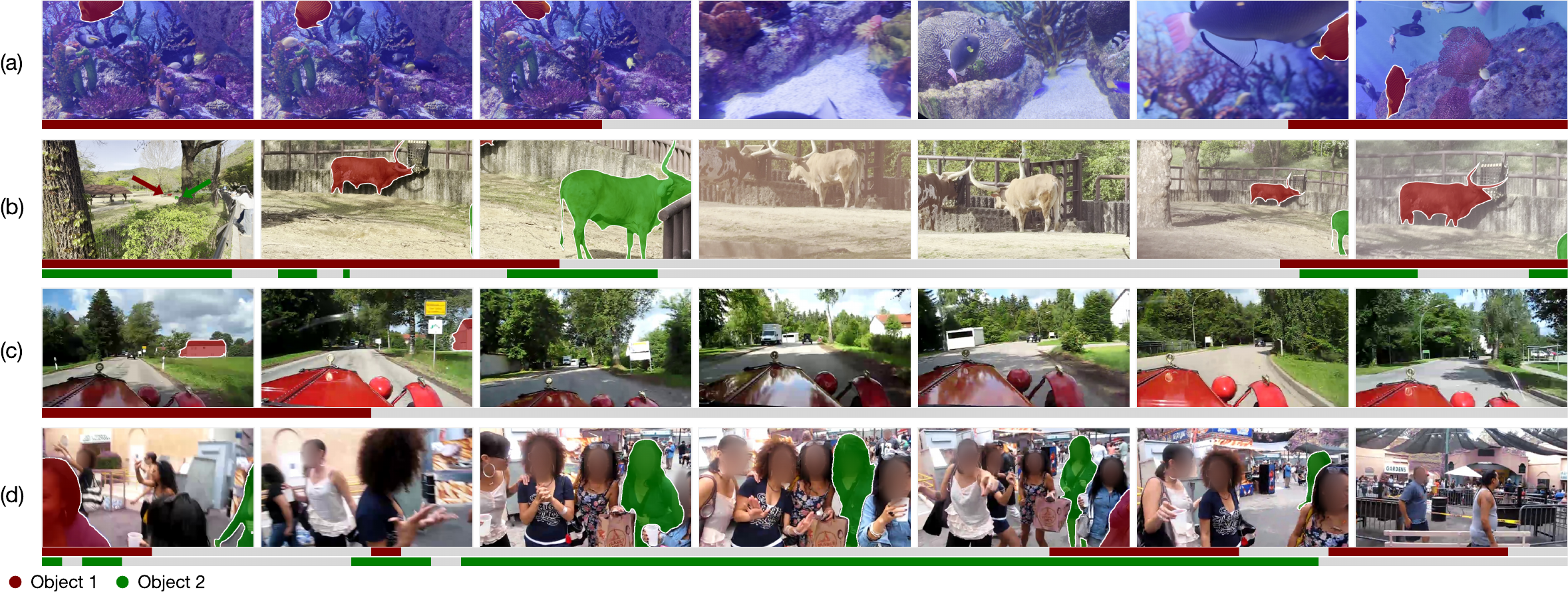}
  \vspace{-20pt}
  \caption{Example videos from the FaVOS benchmark. Each row shows sampled frames from a video with annotated target masks. The bars below each sequence indicate temporal visibility, where colored segments denote frames in which the target object is visible and gray segments denote absence. See Appendix~\ref{app:additional_examples} for additional examples.}
  \label{fig:favos_example}
    \vspace{-10pt}
\end{figure}

\begin{wrapfigure}{r}{0.35\linewidth}
  \vspace{-25pt}
  \centering
  \includegraphics[width=\linewidth]{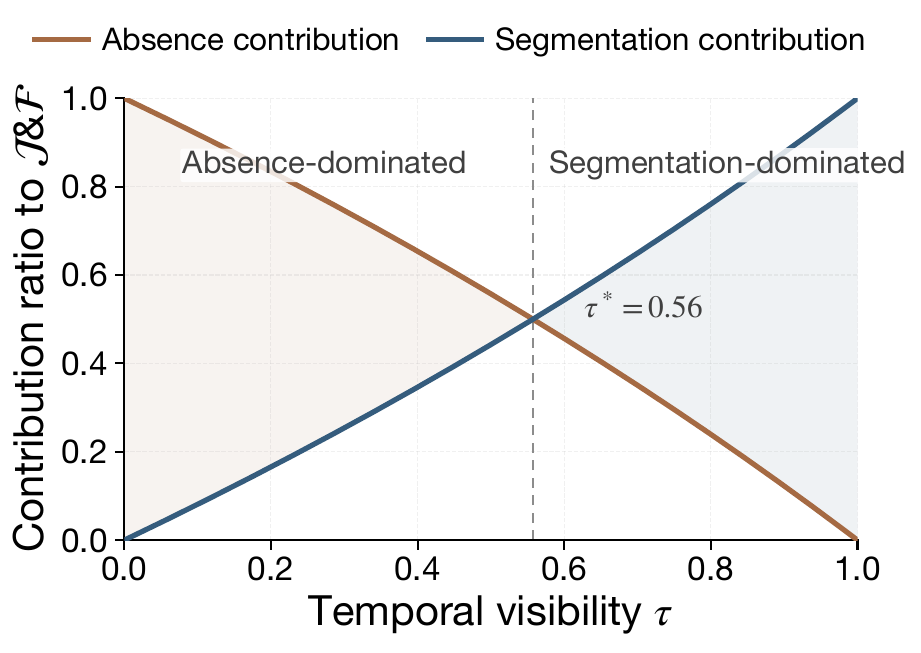}
  \vspace{-20pt}
  \caption{Contribution ratio of classification and segmentation terms in \jnf{} as a function of temporal visibility $\tau$.}
  \label{fig:jnf_contribution_ratio}
  \vspace{-14pt}
\end{wrapfigure}

\section{Volumetric \jnf{}}

\subsection{\jnf{} Collapse under Low Temporal Visibility}
\label{sec.5.1}

We first analyze how \jnf{} behaves when the target object is absent. From Eq.~\ref{eq:iou_definition}, \jnf{} evaluates segmentation quality only when the target object is present. When the ground-truth mask is empty, the frame-level score reduces to
\begin{equation}
    \mathbf{1}\left(\|\hat{\mathcal{M}}\|_0 = 0\right),
\end{equation}
depending solely on whether the prediction is also empty.

Using Eq.~\ref{eq:jnf_decomposition}, the segmentation term is scaled by $\tau p_1$, whereas the target-absence term is scaled by $1-\tau$. As temporal visibility decreases, \jnf{} therefore becomes increasingly dominated by $p_0$, while the influence of segmentation quality $s$ diminishes.

We define the normalized contributions of the absence-classification and segmentation terms as
\begin{equation}
R_{\mathrm{cls}}(\tau)
=
\frac{(1-\tau)p_0}{\mathjnf},
\quad
R_{\mathrm{seg}}(\tau)
=
\frac{\tau p_1 s}{\mathjnf},
\end{equation}
where $R_{\mathrm{cls}}(\tau) + R_{\mathrm{seg}}(\tau) = 1$.

Fig.~\ref{fig:jnf_contribution_ratio} shows that as temporal visibility decreases, $R_{\mathrm{cls}}$ increases while $R_{\mathrm{seg}}$ decreases, indicating that \jnf{} increasingly reflects absence classification rather than segmentation quality. With $p_0=p_1=s=0.8$, the absence-classification contribution exceeds the segmentation contribution when $\tau < 0.56$. Similar trends under other parameter settings are provided in Appendix~\ref{app:jnf_contribution}.

We further validate this phenomenon empirically in Sec.~\ref{sec:experiments}, where even a trivial empty-mask predictor can outperform state-of-the-art methods in low-visibility regimes.

\begin{table}[t]
  \centering
  \setlength{\tabcolsep}{4pt}
  \renewcommand{\arraystretch}{0.95}
  \footnotesize
  \caption{Performance comparison on the FaVOS benchmark.}
  % \vspace{-4pt}
  {\setlength{\arrayrulewidth}{0.3pt}
  \resizebox{1.0\textwidth}{!}{
    \begin{tabular}{l l|
    c!{\vrule width 0.3pt}c!{\vrule width 0.3pt}c c c|
    c!{\vrule width 0.3pt}c!{\vrule width 0.3pt}c c c}
      \toprule
      \textbf{Model} & \textbf{Venue}
      & \multicolumn{5}{c|}{\textbf{FaVOS-20}}
      & \multicolumn{5}{c}{\textbf{FaVOS-40}} \\

      \cmidrule(lr){3-7} \cmidrule(lr){8-12}
      &
      & $\mathcal{J}\&\mathcal{F}$
      & \jnfv{}
      & $p_0$ & $p_1$ & $s$
      & $\mathcal{J}\&\mathcal{F}$
      & \jnfv{}
      & $p_0$ & $p_1$ & $s$ \\
      \midrule

      \rowcolor{blue!8}
      \textbf{Empt. pred.} 
          & -- 
        & 80.0 & 0.0 & 1.00 & 0.00 & 0.00
        & 59.9 & 0.0 & 1.00 & 0.00 & 0.00 \\

    STM~\cite{oh2019stm} & ICCV'19
        & 71.9 & 35.5 & 0.82 & 0.62 & 0.65
        & 65.9 & 44.9 & 0.85 & 0.64 & 0.68 \\
    
    STCN~\cite{cheng2021stcn} & NeurIPS'21
        & 50.6 & 26.0 & 0.52 & 0.81 & 0.60
        & 53.9 & 37.7 & 0.57 & 0.85 & 0.63 \\
    
    XMem~\cite{cheng2022xmem} & ECCV'22
        & 63.8 & 36.8 & 0.69 & 0.74 & 0.66
        & 62.0 & 45.3 & 0.71 & 0.78 & 0.69 \\
    
    DeAOT-L~\cite{yang2022deaot} & NeurIPS'22
        & 59.9 & 38.0 & 0.61 & 0.80 & 0.73
        & 62.7 & 49.2 & 0.65 & 0.83 & 0.77 \\
        
    Cutie-B~\cite{cheng2024cutie} & CVPR'24
        & 80.4 & 49.0 & 0.88 & 0.71 & 0.78
        & 76.7 & 58.6 & 0.94 & 0.71 & 0.84 \\

    % SAM 2.1-B+~\cite{ravi2024sam2} & ICLR'25
    %     & 73.6 & 47.5 & 0.76 & 0.82 & 0.85
    %     & 79.6 & 59.5 & 0.89 & 0.82 & 0.87 \\
    
    SAM 2.1-L~\cite{ravi2024sam2} & ICLR'25
        & 73.4 & 46.7 & 0.76 & 0.81 & 0.85
        & 80.1 & 60.5 & 0.90 & 0.83 & 0.87 \\

    SAMURAI-L~\cite{yang2024samurai} & arXiv'24 
        & 60.8 & 37.3 & 0.58 & 0.91 & 0.82
        & 72.1 & 54.9 & 0.68 & 0.93 & 0.87 \\

    DAM4SAM-L~\cite{videnovic2025didi} & CVPR'25
        & 69.6 & 44.3 & 0.70 & 0.85 & 0.84
        & 78.0 & 63.2 & 0.82 & 0.88 & 0.89 \\

    % SAM2Long-B~\cite{ding2025sam2long} & ICCV'25
    %     & 72.8 & 44.4 & 0.75 & 0.82 & 0.82
    %     & 79.2 & 62.2 & 0.84 & 0.87 & 0.88 \\
    
    SAM2Long-L~\cite{ding2025sam2long} & ICCV'25
        & 74.0 & 49.8 & 0.76 & 0.84 & 0.83
        & 78.3 & 64.8 & 0.83 & 0.87 & 0.87 \\
    
    SeC~\cite{zhang2025secvos} & ICLR'26
        & 80.6 & 55.7 & 0.84 & 0.79 & 0.88
        & 84.0 & 70.9 & 0.91 & 0.85 & 0.92 \\
        
    SAM 3~\cite{carion2025sam3} & ICLR'26
        & 78.7 & 56.0 & 0.80 & 0.89 & 0.84
        & 84.4 & 72.7 & 0.88 & 0.90 & 0.90 \\

      \bottomrule
    \end{tabular}
  }}
  \label{tab:favos-main}
\end{table}

\subsection{Definition of Volumetric \jnf{}}

The collapse above stems from the fact that frame-wise IoU is not naturally defined when both the ground-truth and predicted masks are empty. Standard \jnf{} handles this case with a special target-absence rule, explicitly rewarding empty predictions with \textit{perfect scores} that are substantially harder to achieve when the target is present. We instead reformulate evaluation at the video-volume level. By stacking masks into spatio-temporal volumes, both region and boundary agreement can be evaluated without assigning explicit rewards to target-absent frames. Volumetric \jnf{} (\jnfv{}) is defined as the average of the volumetric Jaccard index $\mathcal{J}_v$ and volumetric contour accuracy $\mathcal{F}_v$.

\paragraph{Volumetric Jaccard.}
We extend the Jaccard index to the spatio-temporal setting by treating a mask sequence as a 3D binary volume. Given ground-truth and predicted mask sequences $\mathcal{M}=\{\mathcal{M}_t\}_{t=1}^{T}$ and $\hat{\mathcal{M}}=\{\hat{\mathcal{M}}_t\}_{t=1}^{T}$, we define the corresponding spatio-temporal volumes $\mathcal{V}$ and $\hat{\mathcal{V}}$ by stacking masks along the temporal axis. The volumetric Jaccard index is defined as
\begin{equation}
    \mathcal{J}_{v}(\mathcal{V},\hat{\mathcal{V}})
    =
    \frac{
        \|\mathcal{V} \cap \hat{\mathcal{V}}\|_0
    }{
        \|\mathcal{V} \cup \hat{\mathcal{V}}\|_0
    }
    =
    \frac{
        \sum_{t=1}^{T}
        \|\mathcal{M}_t \cap \hat{\mathcal{M}}_t\|_0
    }{
        \sum_{t=1}^{T}
        \|\mathcal{M}_t \cup \hat{\mathcal{M}}_t\|_0
    }.
\end{equation}

Equivalently, $\mathcal{J}_{v}$ can be interpreted as a union-volume-weighted average of the per-frame Jaccard index:
\begin{equation}
    \mathcal{J}_{v}
    =
    \frac{
        \sum_{t=1}^{T}
        u_t \,
        \mathcal{J}(\mathcal{M}_t,\hat{\mathcal{M}}_t)
    }{
        \sum_{t=1}^{T}
        u_t
    },
    \qquad
    u_t=\|\mathcal{M}_t \cup \hat{\mathcal{M}}_t\|_0.
    \label{eq:volume_jaccard_unweighted}
\end{equation}

Unlike the standard frame-wise average, $\mathcal{J}_{v}$ does not assign additional reward to frames where both masks are empty. Importantly, target-absent frames are still penalized when the model predicts false-positive masks, since such predictions increase the union volume without increasing the intersection. More generally, weighted temporal aggregation can be written as
\begin{equation}
    \mathcal{J}_{w}
    =
    \frac{
        \sum_{t=1}^{T}
        w(u_t)\,
        \mathcal{J}(\mathcal{M}_t,\hat{\mathcal{M}}_t)
    }{
        \sum_{t=1}^{T}
        w(u_t)
    },
    \label{eq:volume_jaccard_weighted}
\end{equation}
where $w(u_t)=u_t$ recovers $\mathcal{J}_v$, while $w(u_t)=\mathbf{1}(u_t>0)$ yields the vIoU-style aggregation used in Long-RVOS~\cite{liang2026longrvos}. A detailed derivation is provided in Appendix~\ref{app:derivation_volumetric_jaccard}.

\paragraph{Volumetric Contour Accuracy.}
Following the DAVIS~\cite{perazzi2016davis} contour evaluation protocol, we define volumetric contour accuracy by matching boundary voxels between ground-truth and predicted spatio-temporal mask volumes. Let $c(\mathcal{V})$ and $c(\hat{\mathcal{V}})$ denote the corresponding boundary voxel sets, where two voxels are matched if they fall within spatial tolerance $\delta_{xy}$ and temporal tolerance $\delta_t$.

Let $m(\mathcal{V},\hat{\mathcal{V}})$ denote the cardinality of the resulting maximum matching. The volumetric contour precision, recall, and contour accuracy are defined as
\begin{equation}
P_c^{v}
=
\frac{m(\mathcal{V},\hat{\mathcal{V}})}{|c(\hat{\mathcal{V}})|},
\qquad
R_c^{v}
=
\frac{m(\mathcal{V},\hat{\mathcal{V}})}{|c(\mathcal{V})|},
\qquad
\mathcal{F}_{v}
=
\frac{2P_c^{v}R_c^{v}}{P_c^{v}+R_c^{v}}.
\end{equation}

Implementation details and additional configurations are provided in Appendix~\ref{app:volumetric_contour_accuracy}.

\vspace{1.0em}
\section{Experiments}
\label{sec:experiments}

\subsection{Model Behavior under Low Temporal Visibility}

\begin{wrapfigure}{r}{0.45\linewidth}
  \vspace{-1.5em}
  \centering
  \includegraphics[width=\linewidth]{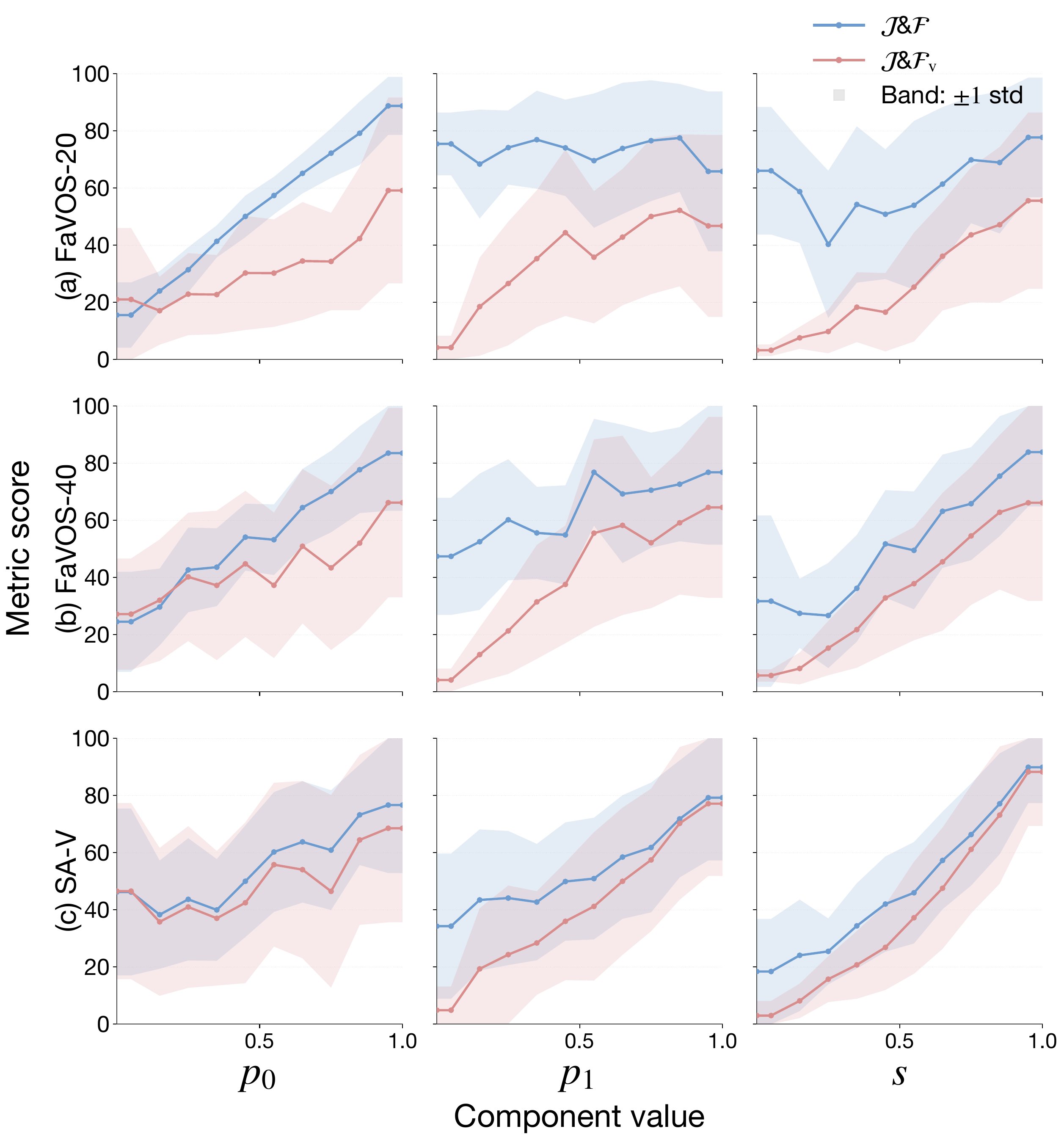}
  % \vspace{-14pt}
  \caption{Object-level dependency of \jnf{} and \jnfv{} on $p_0$, $p_1$, and $s$, 
across (a) FaVOS-20, (b) FaVOS-40, and (c) SA-V.}
  \label{fig:jnf_vs_p0p1s}
  \vspace{-1em}
\end{wrapfigure}

In this section, we analyze how existing VOS models behave under low temporal visibility. Tab.~\ref{tab:favos-main} reports results on \benchname{}-20 and \benchname{}-40 using \jnf{}~\cite{perazzi2016davis}, its decomposition into \pzero{} ($p_0$), \pone{} ($p_1$), and \s{} ($s$), and then volumetric \jnf{} (\jnfv{}).
As temporal visibility decreases, \jnf{} deviates from the trend commonly observed in standard benchmarks, where more recent models achieve higher performance. Instead, \jnf{} becomes increasingly dominated by \pzero{} ($p_0$), while $p_0$ and $p_1$ exhibit a clear trade-off.

Across models, \s{} ($s$) generally improves with model advancement, indicating increasingly accurate segmentation on target-present frames. However, under low temporal visibility, this improvement is not consistently reflected in \jnf{}, which remains strongly influenced by $p_0$.

For instance, Cutie-B~\cite{cheng2024cutie} achieves lower values for both $p_1$ and $s$ than recent SAM2-based variants~\cite{ravi2024sam2,yang2024samurai,videnovic2025didi,ding2025sam2long} and SAM3~\cite{carion2025sam3}, yet attains a higher \jnf{} score on \benchname{}-20 due to its substantially higher $p_0$. Similarly, STM~\cite{oh2019stm} achieves competitive \jnf{} despite relatively weaker segmentation quality. In the extreme case, an empty-mask predictor with $p_1=s=0$ can still match or outperform several state-of-the-art methods solely through perfect \pzero{} ($p_0=1$). These results suggest that, under low temporal visibility, \jnf{} increasingly rewards conservative absence prediction over meaningful segmentation quality.

Notably, tracking-based SAM2 variants (e.g., SAMURAI, DAM4SAM, and SAM2Long) do not substantially improve $s$, but instead shift the $p_0$–$p_1$ trade-off toward higher $p_1$. While beneficial under high temporal visibility, this behavior is less effective in low-visibility settings. Overall, these results show that \jnf{} favors absence accuracy over segmentation quality under low temporal visibility.

\subsection{Behavior of \jnf{} and \jnfv{} under Varying Temporal Visibility}

The behavior of \jnf{} and \jnfv{} under varying temporal visibility is shown in Fig.~\ref{fig:jnf_vs_p0p1s}. 
We construct the curves by binning objects based on $p_0$, $p_1$, and $s$ with a bin width of 0.1, and plotting the mean metric value within each bin across all methods in Tab.~\ref{tab:favos-main}.
In low temporal visibility settings (Fig.~\ref{fig:jnf_vs_p0p1s} (a) FaVOS-20), \jnf{} exhibits a strong monotonic dependence on $p_0$, indicating that its score is largely dominated by absence classification. 
However, this dependence weakens as temporal visibility increases (Fig.~\ref{fig:jnf_vs_p0p1s} (b) FaVOS-40 and (c) SA-V).
In contrast, \jnfv{} shows substantially reduced sensitivity to $p_0$ even under low temporal visibility, and remains largely consistent across different visibility regimes.

For $p_1$ and $s$, \jnf{} remains largely insensitive under low temporal visibility, with nearly flat trends, suggesting that \pone{} and \s{} have limited influence on the final score. 
This confirms that the evaluation effectively collapses to an absence classification task. 
On the other hand, \jnfv{} consistently exhibits increasing trends with respect to both $p_1$ and $s$, even under low visibility, \rev{indicating that its sensitivity to these components remains stable across temporal-visibility regimes.}

\begin{table}[t]
  \centering
  \setlength{\tabcolsep}{3.6pt}
  \renewcommand{\arraystretch}{0.95}
  \scriptsize
  \caption{Performance comparison on FaVOS-full and SA-V~\cite{ravi2024sam2} test split.}
  {\setlength{\arrayrulewidth}{0.3pt}
  \resizebox{\textwidth}{!}{
    \begin{tabular}{l|
    c c c c c c|
    c c c c c c}
      \toprule
      \textbf{Model}
      & \multicolumn{6}{c|}{\textbf{FaVOS-full}}
      & \multicolumn{6}{c}{\textbf{SA-V test}} \\
      \cmidrule(lr){2-7} \cmidrule(lr){8-13}
& \jnf{}$\uparrow$ 
& \jnfv{}$\uparrow$ 
& $\mathrm{tIoU}\uparrow$ 
& $\mathcal{T}\downarrow$ 
& $\mathcal{J}\&\dot{\mathcal{F}}_d\uparrow$ 
& $\mathcal{J}\&\dot{\mathcal{F}}_r\uparrow$
& \jnf{}$\uparrow$ 
& \jnfv{}$\uparrow$ 
& $\mathrm{tIoU}\uparrow$ 
& $\mathcal{T}\downarrow$ 
& $\mathcal{J}\&\dot{\mathcal{F}}_d\uparrow$ 
& $\mathcal{J}\&\dot{\mathcal{F}}_r\uparrow$ \\
      \midrule

      \rowcolor{blue!8}
      \textbf{Empt. pred.}
        & 69.2 & 0.0 & 0.00 & -- & 100.0 & 0.0 
        & 25.3 & 0.0 & 0.00 & -- & 100.0 & 0.0  \\

      STM~\cite{oh2019stm}
        & 68.6 & 40.5 & 0.45 & 0.57 & 82.3 & 16.3
        & 46.6 & 34.4 & 0.50 & 0.56 & 82.9 & 21.1 \\

      STCN~\cite{cheng2021stcn}
        & 52.4 & 32.3 & 0.40 & 0.56 & 54.4 & 36.4
        & 60.4 & 55.1 & 0.70 & 0.51 & 70.8 & 43.5 \\

      XMem~\cite{cheng2022xmem}
        & 62.8 & 41.4 & 0.47 & 0.56 & 68.8 & 28.5
        & 58.0 & 48.9 & 0.63 & 0.52 & 81.6 & 28.4 \\

      DeAOT-L~\cite{yang2022deaot}
        & 61.4 & 44.0 & 0.44 & 0.54 & 61.2 & 35.7
        & 62.2 & 55.9 & 0.75 & 0.51 & 63.2 & 44.7 \\

      Cutie-B~\cite{cheng2024cutie}
        & 78.4 & 54.2 & 0.58 & 0.55 & 90.3 & 26.1
        & 62.2 & 53.4 & 0.64 & 0.53 & 86.8 & 31.5 \\

      SAM 2.1-L~\cite{ravi2024sam2}
        & 77.0 & 54.1 & 0.56 & 0.53 & 83.6 & 41.1
        & 79.7 & 74.3 & 0.84 & 0.51 & 84.3 & 59.3 \\

      SAMURAI-L~\cite{yang2024samurai}
        & 66.9 & 46.7 & 0.50 & 0.53 & 65.1 & 57.6
        & 79.9 & 76.4 & 0.89 & 0.51 & 71.2 & 63.9 \\

      DAM4SAM-L~\cite{videnovic2025didi}
        & 74.1 & 54.5 & 0.56 & 0.52 & 77.3 & 48.4
        & 80.1 & 76.6 & 0.87 & 0.50 & 85.7 & 60.8 \\

      SAM2Long-L~\cite{ding2025sam2long}
        & 76.3 & 57.9 & 0.61 & 0.52 & 80.6 & 45.8
        & 80.4 & 76.4 & 0.87 & 0.50 & 82.9 & 59.8 \\

      SeC~\cite{zhang2025secvos}
        & 82.4 & 63.9 & 0.64 & 0.53 & 88.2 & 48.4
        & 81.7 & 77.8 & 0.88 & 0.50 & 86.3 & 58.9 \\
        
      SAM 3~\cite{carion2025sam3}
        & 81.8 & 65.0 & 0.66 & 0.53 & 83.0 & 58.8
        & 84.7 & 82.6 & 0.91 & 0.51 & 83.0 & 68.4 \\

      \bottomrule
    \end{tabular}
  }}
  \label{tab:alternative_metrics}
\end{table}

\subsection{Comparison with Alternative Evaluation Metrics}

Tab.~\ref{tab:alternative_metrics} compares \jnf{} and \jnfv{} with existing temporal metrics, including tIoU~\cite{liang2026longrvos}, $\mathcal{T}$~\cite{perazzi2016davis}, $\mathcal{J}\&\dot{\mathcal{F}}_d$, and $\mathcal{J}\&\dot{\mathcal{F}}_r$~\cite{mosev2}. On SA-V~\cite{ravi2024sam2}, both \jnf{} and \jnfv{} generally increase with model performance, and the collapse observed under low temporal visibility is much less pronounced. 
This suggests that the failure mode mainly arises in low-visibility regimes; under high temporal visibility, \jnf{} remains largely aligned with \jnfv{}, with no substantial re-ranking. \rev{This agreement is also observed on DAVIS17~\cite{perazzi2016davis} and OVOS~\cite{miao2024ovos}, where the model rankings are largely preserved (Appendix~\ref{app:backward_compat}).}

tIoU follows the same overall trend and serves as a useful indicator of temporal correctness, but it does not fully explain segmentation performance. For example, methods such as STM~\cite{oh2019stm} achieve competitive \jnf{} despite relatively low tIoU, outperforming models with higher temporal overlap. The temporal instability metric $\mathcal{T}$ steadily decreases for stronger models, although the absolute differences remain small.

\begin{wrapfigure}{r}{0.4\linewidth}
  % \vspace{-1.5em}
  \centering
  \includegraphics[width=\linewidth]{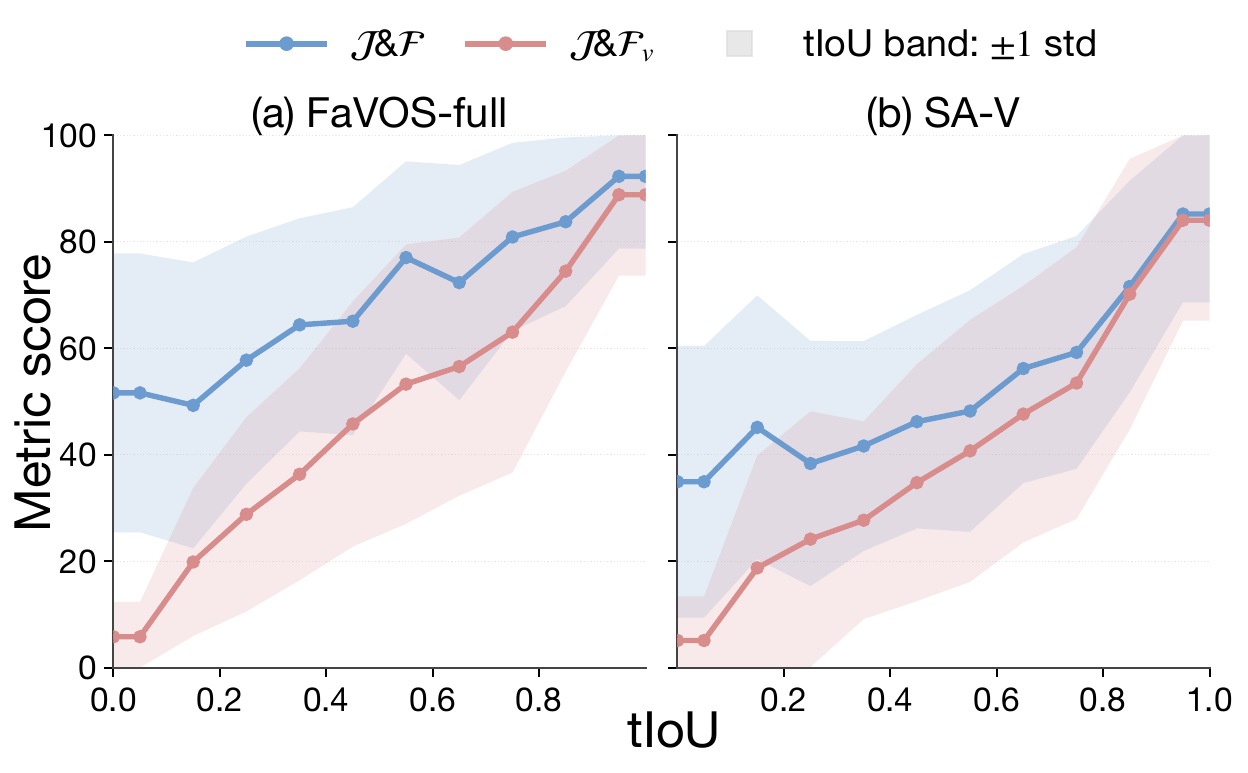}
  \vspace{-1.em}
  \caption{Relationship between temporal overlap (tIoU) and segmentation metrics on (a) \benchname{}-full and (b) SA-V test.}
  \label{fig:tiou_relation}
  % \vspace{-1.0em}
\end{wrapfigure}

The decomposed metrics $\mathcal{J}\&\dot{\mathcal{F}}_d$ and $\mathcal{J}\&\dot{\mathcal{F}}_r$ exhibit a trade-off similar to that of $p_0$ and $p_1$. In particular, $\mathcal{J}\&\dot{\mathcal{F}}_d$ behaves similarly to \pzero{} and can favor models with high $p_0$, including degenerate empty-mask predictors, while being vulnerable to models with low $p_0$. In contrast, $\mathcal{J}\&\dot{\mathcal{F}}_r$ tends to follow $p_1$, although not strictly. These metrics are therefore useful as supplementary diagnostics, but are less suitable as standalone primary metrics. \rev{Similarly, a visibility-balanced \jnf{} still assigns 50.0 to the empty-mask predictor, and a two-number protocol of present-frame \jnf{} and presence F1 does not yield a single model ranking (Appendix~\ref{app:alternative_protocols}).}

Furthermore, Fig.~\ref{fig:tiou_relation} shows that \jnfv{} better reflects tIoU than \jnf{} on both \benchname{}-full and SA-V, covering both low- and high-temporal-visibility regimes. In particular, \jnf{} is less sensitive to temporal overlap when tIoU is low, whereas \jnfv{} varies more consistently with tIoU. This indicates that \jnfv{} better captures temporal properties of video segmentation while preserving the region-and-contour structure of \jnf{}.

\begin{revblock}
\subsection{Robustness to Frame Weighting}
\label{sec:weighting}

\begin{table}[t]
\begin{revblock}
  \centering
  \setlength{\tabcolsep}{3.3pt}
  \renewcommand{\arraystretch}{1.05}
  \footnotesize
  \caption{\jnfv{} under different frame weightings $w(u_t)$. Arrows mark rank changes from the shaded linear weighting used in our main experiments, computed from unrounded scores. Rows are sorted by the linear-weighting score on \benchname{}-20.}
  {\setlength{\arrayrulewidth}{0.3pt}
  \begin{tabular}{l|>{\columncolor{gray!7}}cccc|>{\columncolor{gray!7}}cccc|>{\columncolor{gray!7}}cccc}
    \toprule
    & \multicolumn{4}{c|}{\textbf{\benchname{}-20}} & \multicolumn{4}{c|}{\textbf{\benchname{}-40}} & \multicolumn{4}{c}{\textbf{SA-V test}} \\
    \cmidrule(lr){2-5} \cmidrule(lr){6-9} \cmidrule(lr){10-13}
    \textbf{Model} & Linear & Sqrt & Log & Uniform & Linear & Sqrt & Log & Uniform & Linear & Sqrt & Log & Uniform \\
    \midrule
    SAM 3 & 56.0 & 55.5 & 55.8 & 54.5 & 72.7 & 71.7 & 72.1 & 70.0 & 82.6 & 82.3 & 82.5 & 81.2 \\
    SeC & 55.7 & 55.2 & 55.4 & 54.1 & 70.9 & 70.5 & 70.8 & 69.4 & 77.8 & 77.6 & 77.8 & 76.7 \\
    SAM2Long-L & 49.8 & 49.4 & 49.7 & 48.5 & 64.8 & 64.5 & 64.7 & 63.7 & 76.4 & 76.3 & 76.5 & 75.5 \\
    Cutie-B & 49.0 & 48.5 & 48.7 & 47.2 & 58.6 & 58.4 & 58.6 & 57.6 & 53.4 & 53.2 & 53.4 & 52.6 \\
    SAM 2.1-L & 46.7 & 45.7 & 46.0 & 44.2 & 60.5 & 60.3 & 60.5 & 59.6 & 74.3 & 74.4 & 74.5 & 73.8 \\
    DAM4SAM-L & 44.3 & 44.2 & 44.4 & 43.5 & 63.2 & 62.8 & 63.0 & 61.6 & 76.6 & 76.2$^{\downarrow2}$ & 76.5$^{\downarrow2}$ & 75.1$^{\downarrow2}$ \\
    DeAOT-L & 38.0 & 37.0$^{\downarrow1}$ & 37.2$^{\downarrow1}$ & 34.1$^{\downarrow1}$ & 49.2 & 48.6 & 48.9 & 45.7 & 55.9 & 55.8 & 55.9 & 54.9 \\
    SAMURAI-L & 37.3 & 37.3$^{\uparrow1}$ & 37.3$^{\uparrow1}$ & 36.9$^{\uparrow1}$ & 54.9 & 55.0 & 55.1 & 54.2 & 76.4 & 76.6$^{\uparrow2}$ & 76.6$^{\uparrow2}$ & 76.0$^{\uparrow2}$ \\
    XMem & 36.8 & 36.0 & 36.2 & 33.8 & 45.3 & 44.7 & 45.0 & 42.7 & 48.9 & 48.8 & 48.8 & 48.1 \\
    STM & 35.5 & 34.8 & 35.1 & 32.4 & 44.9 & 44.0 & 44.5 & 41.2 & 34.4 & 33.8 & 34.0 & 32.9 \\
    STCN & 26.0 & 25.5 & 25.8 & 24.2 & 37.7 & 37.0 & 37.4 & 34.8 & 55.1 & 54.8 & 55.0 & 53.6 \\
    \rowcolor{blue!8}
    \textbf{Empt. pred.} & 0.0 & 0.0 & 0.0 & 0.0 & 0.0 & 0.0 & 0.0 & 0.0 & 0.0 & 0.0 & 0.0 & 0.0 \\
    \bottomrule
  \end{tabular}}
  \label{tab:weighting}
\end{revblock}
\end{table}

Since $\mathcal{J}_v$ is a union-volume-weighted average of frame-wise IoU (Eq.~\ref{eq:volume_jaccard_unweighted}), frames with larger object regions have greater influence on \jnfv{}. To examine whether our conclusions depend on this choice, we recompute \jnfv{} for all models by replacing the linear weighting $w(u_t)=u_t$ in Eq.~\ref{eq:volume_jaccard_weighted} with square-root weighting $w(u_t)=\sqrt{u_t}$, a normalized logarithmic weighting relative to the median non-zero union size of each object sequence, and uniform weighting $w(u_t)=\mathbf{1}(u_t>0)$, which assigns equal weight to every frame with a non-empty union.

As shown in Tab.~\ref{tab:weighting}, the model rankings are highly robust to the weighting function. The \benchname{}-40 rankings are unchanged under every weighting. On \benchname{}-20, the only change is a swap between DeAOT-L and SAMURAI-L, whose linear-weighting scores differ by less than one point. On SA-V, DAM4SAM-L and SAMURAI-L swap ranks 3 and 5, and the three methods involved lie within 0.2 points of each other under linear weighting. SAM~3 and SeC remain the top two methods, and the empty-mask predictor receives zero under every weighting.
\end{revblock}

\begin{revblock}
\section{Limitations}
\label{sec:limitations}

With a non-uniform frame weighting $w(u_t)$, such as the default linear weighting, \jnfv{} can be influenced by object size. The weighting in Eq.~\ref{eq:volume_jaccard_weighted} can be adjusted to the downstream task, and the model rankings remain stable across the alternatives we examine (Sec.~\ref{sec:weighting}). In addition, \jnfv{} penalizes false positives on target-absent frames by their volume rather than per event, so many small hallucinations may incur only a mild penalty. Applications in which each false alarm carries a separate cost should report an event-level measure alongside \jnfv{}.

\benchname{} contains 200 videos with \corr{279} objects, fewer than large-scale datasets such as MOSEv2~\cite{mosev2} and Long-RVOS~\cite{liang2026longrvos}, which also serve as training data. \benchname{} is instead designed as a diagnostic evaluation benchmark that isolates the low-temporal-visibility regime, and its scale is comparable to widely used evaluation sets such as DAVIS17~\cite{perazzi2016davis} and the SA-V test split~\cite{ravi2024sam2}. Expanding its scenario coverage and diversity remains future work.
\end{revblock}

\section{Conclusion}
In this work, we revisit the standard evaluation protocol for video object segmentation under low temporal visibility. We show that, in this regime, the conventional \jnf{} metric can become dominated by absence classification accuracy, effectively collapsing into a reward for predicting target absence. We formalize this behavior theoretically and introduce \textbf{\benchname{}}, a benchmark designed to evaluate VOS under controlled low-visibility scenarios. Through empirical analysis, we show that \jnf{} can substantially favor models with high \pzero{}, even when their \pone{} and \s{} are limited.

To address this limitation, we introduce \textbf{volumetric \jnf{} (\jnfv{})}, which evaluates predicted masks as spatio-temporal volumes rather than independent frame-wise masks. Extensive experiments show that \jnfv{} reduces the dominance of target-absence rewards while remaining sensitive to true positive rate, segmentation quality, and temporal consistency. We further show that \jnfv{} better reflects temporal consistency, making it a more suitable evaluation criterion for long and complex videos. We hope that \benchname{} and \jnfv{} encourage future work on VOS methods that move beyond absence classification and achieve reliable pixel-level understanding over time.

\newpage
\renewcommand{\acksection}{\section*{Acknowledgements}}
\begin{ack}
\corr{This work was supported in part by the National Research Foundation of Korea (NRF) grants (RS-2025-00560762, RS-2024-00414981), the Institute of Information \& Communications Technology Planning \& Evaluation (IITP) grants (RS-2025-25442338, RS-2026-25518317, RS-2026-25551943, RS-2021-II211343), and the Ministry of Science and ICT under the Advanced GPU Utilization Support Program. J. Do is with ASRI, Seoul National University.}
\end{ack}

\bibliographystyle{plain}
\bibliography{main}

@String(IJCV  = {IJCV})

@String(CVPR  = {IEEE/CVF Conference on Computer Vision and Pattern Recognition (CVPR)})

@String(ICCV  = {IEEE/CVF International Conference on Computer Vision (ICCV)})

@String(ECCV  = {European Conference on Computer Vision (ECCV)})

@String(NIPS  = {Conference on Neural Information Processing Systems (NeurIPS)})

@String(NeurIPS  = {Conference on Neural Information Processing Systems (NeurIPS)})

@String(TPAMI  = {IEEE Transactions on Pattern Analysis and Machine Intelligence (TPAMI)})

@String(IJCV  = {International Journal of Computer Vision (IJCV)})

@String(TIP   = {IEEE TIP})

@String(ACCV  = {Asian Conference on Computer Vision (ACCV)})

@String(ICLR  = {International Conference on Learning Representations (ICLR)})

@String(AAAI = {AAAI Conference on Artificial Intelligence (AAAI)})

@inproceedings{cheng2021stcn,
  title={{Rethinking Space-Time Networks with Improved Memory Coverage for Efficient Video Object Segmentation}},
  author={Cheng, Ho Kei and Tai, Yu-Wing and Tang, Chi-Keung},
  booktitle=NIPS,
  year={2021}
}

@inproceedings{khoreva2018refdavis,
  title={{Video Object Segmentation with Language Referring Expressions}},
  author={Khoreva, Anna and Rohrbach, Anna and Schiele, Bernt},
  booktitle=ACCV,
  year={2018}
}

@inproceedings{zhang2025secvos,
  title={{Advancing Complex Video Object Segmentation via Progressive Concept Construction}},
  author={Zhang, Zhixiong and Ding, Shuangrui and Dong, Xiaoyi and He, Songxin and Lin, Jianfan and Tang, Junsong and Zang, Yuhang and Cao, Yuhang and Lin, Dahua and Wang, Jiaqi},
  booktitle=ICLR,
  year={2026}
}

@article{mosev2,
  title={{MOSEv2: A More Challenging Dataset for Video Object Segmentation in Complex Scenes}},
  author={Ding, Henghui and Ying, Kaining and Liu, Chang and He, Shuting and Jiang, Xudong and Jiang, Yu-Gang and Torr, Philip HS and Bai, Song},
  journal={arXiv preprint arXiv:2508.05630},
  year={2025}
}

@inproceedings{mose,
  title={{MOSE: A New Dataset for Video Object Segmentation in Complex Scenes}},
  author={Ding, Henghui and Liu, Chang and He, Shuting and Jiang, Xudong and Torr, Philip HS and Bai, Song},
  booktitle=ICCV,
  year={2023}
}

@inproceedings{ravi2024sam2,
  title={{SAM 2: Segment Anything in Images and Videos}},
  author={Ravi, Nikhila and Gabeur, Valentin and Hu, Yuan-Ting and Hu, Ronghang and Ryali, Chaitanya and Ma, Tengyu and Khedr, Haitham and R{\"a}dle, Roman and Rolland, Chloe and Gustafson, Laura and others},
  booktitle=ICLR,
  year={2025}
}

@article{hong2025lvosv2,
    author={Hong, Lingyi and Liu, Zhongying and Chen, Wenchao and Tan, Chenzhi and Feng, Yuang and Zhou, Xinyu and Guo, Pinxue and Li, Jinglun and Chen, Zhaoyu and Gao, Shuyong and others},
    title={{LVOS: A Benchmark for Large-scale Long-term Video Object Segmentation}},
    journal=TPAMI,
    year = {2025},
}

@inproceedings{hong2023lvosv1,
    author = {Hong, Lingyi and Chen, Wenchao and Liu, Zhongying and Zhang, Wei and Guo, Pinxue and Chen, Zhaoyu and Zhang, Wenqiang},
    title = {{LVOS: A Benchmark for Long-term Video Object Segmentation}},
    booktitle = ICCV,
    year = {2023},
}

@inproceedings{perazzi2016davis,
  title={{A Benchmark Dataset and Evaluation Methodology for Video Object Segmentation}},
  author={Perazzi, Federico and Pont-Tuset, Jordi and McWilliams, Brian and Van Gool, Luc and Gross, Markus and Sorkine-Hornung, Alexander},
  booktitle=CVPR,
  year={2016}
}

@inproceedings{liang2026longrvos,
  title={{Long-RVOS: A Comprehensive Benchmark for Long-term Referring Video Object Segmentation}},
  author={Liang, Tianming and Jiang, Haichao and Yang, Yuting and Tan, Chaolei and Li, Shuai and Zheng, Wei-Shi and Hu, Jian-Fang},
  booktitle=CVPR,
  year={2026}
}

@inproceedings{videnovic2025didi,
  title={{A Distractor-Aware Memory for Visual Object Tracking with SAM2}},
  author={Videnovic, Jovana and Lukezic, Alan and Kristan, Matej},
  booktitle=CVPR,
  year={2025}
}

@inproceedings{kristan2022vot2022,
  title={{The Tenth Visual Object Tracking VOT2022 Challenge Results}},
  author={Kristan, Matej and Leonardis, Ale{\v{s}} and Matas, Ji{\v{r}}{\'\i} and Felsberg, Michael and Pflugfelder, Roman and K{\"a}m{\"a}r{\"a}inen, Joni-Kristian and Chang, Hyung Jin and Danelljan, Martin and Zajc, Luka {\v{C}}ehovin and Luke{\v{z}}i{\v{c}}, Alan and others},
  booktitle={ECCV Workshops},
  year={2022},
}

@inproceedings{fan2019lasot,
  title={{LaSOT: A High-Quality Benchmark for Large-Scale Single Object Tracking}},
  author={Fan, Heng and Lin, Liting and Yang, Fan and Chu, Peng and Deng, Ge and Yu, Sijia and Bai, Hexin and Xu, Yong and Liao, Chunyuan and Ling, Haibin},
  booktitle=CVPR,
  year={2019}
}

@article{huang2019got10k,
  title={{GOT-10k: A Large High-Diversity Benchmark for Generic Object Tracking in the Wild}},
  author={Huang, Lianghua and Zhao, Xin and Huang, Kaiqi},
  journal=TPAMI,
  year={2019}
}

@inproceedings{xu2018ytvos,
  title={{YouTube-VOS: Sequence-to-Sequence Video Object Segmentation}},
  author={Xu, Ning and Yang, Linjie and Fan, Yuchen and Yang, Jianchao and Yue, Dingcheng and Liang, Yuchen and Price, Brian and Cohen, Scott and Huang, Thomas},
  booktitle=ECCV,
  year={2018}
}

@inproceedings{tokmakov2023vost,
  title={{Breaking the ``Object'' in Video Object Segmentation}},
  author={Tokmakov, Pavel and Li, Jie and Gaidon, Adrien},
  booktitle=CVPR,
  year={2023}
}

@inproceedings{ding2023mevis,
  title={{MeViS: A Large-scale Benchmark for Video Segmentation with Motion Expressions}},
  author={Ding, Henghui and Liu, Chang and He, Shuting and Jiang, Xudong and Loy, Chen Change},
  booktitle=ICCV,
  year={2023}
}

@article{ding2025mevisv2,
  title={{MeViS: A Multi-Modal Dataset for Referring Motion Expression Video Segmentation}},
  author={Ding, Henghui and Liu, Chang and He, Shuting and Ying, Kaining and Jiang, Xudong and Loy, Chen Change and Jiang, Yu-Gang},
  journal=TPAMI,
  year={2025}
}

@article{carion2025sam3,
  title={{SAM 3: Segment Anything with Concepts}},
  author={Carion, Nicolas and Gustafson, Laura and Hu, Yuan-Ting and Debnath, Shoubhik and Hu, Ronghang and Suris, Didac and Ryali, Chaitanya and Alwala, Kalyan Vasudev and Khedr, Haitham and Huang, Andrew and others},
  journal=ICLR,
  year={2026}
}

@inproceedings{cheng2022xmem,
  title={{XMem: Long-Term Video Object Segmentation with an Atkinson-Shiffrin Memory Model}},
  author={Cheng, Ho Kei and Schwing, Alexander G},
  booktitle=ECCV,
  year={2022}
}

@inproceedings{cheng2024cutie,
  title={{Putting the Object Back into Video Object Segmentation}},
  author={Cheng, Ho Kei and Oh, Seoung Wug and Price, Brian and Lee, Joon-Young and Schwing, Alexander},
  booktitle=CVPR,
  year={2024}
}

@inproceedings{yang2022deaot,
  title={{Decoupling Features in Hierarchical Propagation for Video Object Segmentation}},
  author={Yang, Zongxin and Yang, Yi},
  booktitle=NeurIPS,
  year={2022}
}

@article{kim2026dpseg,
    title={{Dual Path Semantic Segmentation with Locality Guidance Tokens}},
  author={Kim, Jaeik and Baek, Yoonjae and Jung, Aecheon and Kim, Hongyeob and Choi, Soyun and Kim, Youngeun and Do, Jaeyoung and Hong, Sungeun},
  journal={Available at SSRN 6380741},
    year={2026}
}

@inproceedings{yang2019vis,
  title={{Video Instance Segmentation}},
  author={Yang, Linjie and Fan, Yuchen and Xu, Ning},
  booktitle=ICCV,
  year={2019}
}

@inproceedings{ding2025sam2long,
  title={{SAM2Long: Enhancing SAM 2 for Long Video Segmentation with a Training-Free Memory Tree}},
  author={Ding, Shuangrui and Qian, Rui and Dong, Xiaoyi and Zhang, Pan and Zang, Yuhang and Cao, Yuhang and Guo, Yuwei and Lin, Dahua and Wang, Jiaqi},
  booktitle=ICCV,
  year={2025}
}

@inproceedings{oh2019stm,
  title={{Video Object Segmentation Using Space-Time Memory Networks}},
  author={Oh, Seoung Wug and Lee, Joon-Young and Xu, Ning and Kim, Seon Joo},
  booktitle=ICCV,
  year={2019}
}

@article{thomee2016yfcc100m,
  title={{YFCC100M: The New Data in Multimedia Research}},
  author={Thomee, Bart and Shamma, David A and Friedland, Gerald and Elizalde, Benjamin and Ni, Karl and Poland, Douglas and Borth, Damian and Li, Li-Jia},
  journal={Communications of the ACM (CACM)},
  year={2016}
}

@article{everingham2010pascalvoc,
  title={{The Pascal Visual Object Classes (VOC) Challenge}},
  author={Everingham, Mark and Van Gool, Luc and Williams, Christopher KI and Winn, John and Zisserman, Andrew},
  journal=IJCV,
  year={2010}
}

@article{yao2020vos_survey,
  title={{Video Object Segmentation and Tracking: A Survey}},
  author={Yao, Rui and Lin, Guosheng and Xia, Shixiong and Zhao, Jiaqi and Zhou, Yong},
  journal={ACM Transactions on Intelligent Systems and Technology (TIST)},
  year={2020},
}

@inproceedings{siam2021vcas_driving,
  title={{Video Class Agnostic Segmentation Benchmark for Autonomous Driving}},
  author={Siam, Mennatullah and Kendall, Alex and Jagersand, Martin},
  booktitle=CVPR,
  year={2021}
}

@article{ji2022videopolyp,
  title={{Video Polyp Segmentation: A Deep Learning Perspective}},
  author={Ji, Ge-Peng and Xiao, Guobao and Chou, Yu-Cheng and Fan, Deng-Ping and Zhao, Kai and Chen, Geng and Van Gool, Luc},
  journal={Machine Intelligence Research (MIR)},
  year={2022}
}

@article{zhu2024medsam2,
  title={{Medical SAM 2: Segment medical images as video via Segment Anything Model 2}},
  author={Zhu, Jiayuan and Hamdi, Abdullah and Qi, Yunli and Jin, Yueming and Wu, Junde},
  journal={arXiv preprint arXiv:2408.00874},
  year={2024}
}

@article{patil2021surveillance,
  title={{An Unified Recurrent Video Object Segmentation Framework for Various Surveillance Environments}},
  author={Patil, Prashant W and Dudhane, Akshay and Kulkarni, Ashutosh and Murala, Subrahmanyam and Gonde, Anil Balaji and Gupta, Sunil},
  journal={IEEE Transactions on Image Processing (TIP)},
  year={2021},
}

@article{gao2023vos_survey2,
  title={{Deep learning for video object segmentation: a review}},
  author={Gao, Mingqi and Zheng, Feng and Yu, James JQ and Shan, Caifeng and Ding, Guiguang and Han, Jungong},
  journal={Artificial Intelligence Review},
  year={2023},
}

@inproceedings{gavrilyuk2018a2dsentences,
  title={{Actor and Action Video Segmentation from a Sentence}},
  author={Gavrilyuk, Kirill and Ghodrati, Amir and Li, Zhenyang and Snoek, Cees GM},
  booktitle=CVPR,
  year={2018}
}

@inproceedings{caelles2017osvos,
  title={{One-Shot Video Object Segmentation}},
  author={Caelles, Sergi and Maninis, Kevis-Kokitsi and Pont-Tuset, Jordi and Leal-Taix{\'e}, Laura and Cremers, Daniel and Van Gool, Luc},
  booktitle=CVPR,
  year={2017}
}

@inproceedings{zhou2022avos,
  title={{Audio-Visual Segmentation}},
  author={Zhou, Jinxing and Wang, Jianyuan and Zhang, Jiayi and Sun, Weixuan and Zhang, Jing and Birchfield, Stan and Guo, Dan and Kong, Lingpeng and Wang, Meng and Zhong, Yiran},
  booktitle=ECCV,
  year={2022},
}

@inproceedings{hu2026cutvos,
  title={{Segment Anything Across Shots: A Method and Benchmark}},
  author={Hu, Hengrui and Ying, Kaining and Ding, Henghui},
  booktitle=AAAI,
  year={2026}
}

@inproceedings{wei2022youmvos,
  title={{YouMVOS: An Actor-centric Multi-shot Video Object Segmentation Dataset}},
  author={Wei, Donglai and Kharbanda, Siddhant and Arora, Sarthak and Roy, Roshan and Jain, Nishant and Palrecha, Akash and Shah, Tanav and Mathur, Shray and Mathur, Ritik and Kemkar, Abhijay and others},
  booktitle=CVPR,
  year={2022}
}

@article{yang2024samurai,
  title={{SAMURAI: Adapting Segment Anything Model for Zero-Shot Visual Tracking with Motion-Aware Memory}},
  author={Yang, Cheng-Yen and Huang, Hsiang-Wei and Chai, Wenhao and Jiang, Zhongyu and Hwang, Jenq-Neng},
  journal={arXiv preprint arXiv:2411.11922},
  year={2024}
}

@inproceedings{chen2025m3vos,
  title={{M$^3$-VOS: Multi-Phase, Multi-Transition, and Multi-Scenery Video Object Segmentation}},
  author={Chen, Zixuan and Li, Jiaxin and Liang, Junxuan and Tan, Liming and Guo, Yejie and Lu, Cewu and Li, Yong-Lu},
  booktitle=CVPR,
  year={2025}
}

@article{caelles2019uvos,
  title={{The 2019 DAVIS Challenge on VOS: Unsupervised Multi-Object Segmentation}},
  author={Caelles, Sergi and Pont-Tuset, Jordi and Perazzi, Federico and Montes, Alberto and Maninis, Kevis-Kokitsi and Van Gool, Luc},
  journal={arXiv preprint arXiv:1905.00737},
  year={2019}
}

@inproceedings{pan2022wnet,
  title={{Wnet: Audio-Guided Video Object Segmentation via Wavelet-Based Cross-Modal Denoising Networks}},
  author={Pan, Wenwen and Shi, Haonan and Zhao, Zhou and Zhu, Jieming and He, Xiuqiang and Pan, Zhigeng and Gao, Lianli and Yu, Jun and Wu, Fei and Tian, Qi},
  booktitle=CVPR,
  year={2022}
}

@article{shape2002scd,
  title={{Shape matching and object recognition using shape contexts}},
  author={Belongie, Serge and Malik, Jitendra and Puzicha, Jan},
  journal=TPAMI,
  year={2002},
}

@book{speech1993dtw,
  title={{Fundamentals of Speech Recognition}},
  author={Rabiner, Lawrence and Juang, Biing-Hwang},
  year={1993},
  publisher={Prentice-Hall, Inc.}
}

@inproceedings{wasmuht2026sa,
  title={{The SA-FARI Dataset: Segment Anything in Footage of Animals for Recognition and Identification}},
  author={Wasmuht, Dante and Brookes, Otto and Schall, Maximilian and Palencia, Pablo and Beirne, Christopher and Burghardt, Tilo and Mirmehdi, Majid and K{\"u}hl, Hjalmar and Arandjelovic, Mimi and Pottie, Sam and others},
  booktitle=CVPR,
  year={2026}
}

@article{miao2024ovos,
  title={{Region Aware Video Object Segmentation With Deep Motion Modeling}},
  author={Miao, Bo and Bennamoun, Mohammed and Gao, Yongsheng and Mian, Ajmal},
  journal={IEEE Transactions on Image Processing},
  year={2024},
}

@article{darkhalil2022visor,
  title={{EPIC-KITCHENS VISOR Benchmark: VIdeo Segmentations and Object Relations}},
  author={Darkhalil, Ahmad and Shan, Dandan and Zhu, Bin and Ma, Jian and Kar, Amlan and Higgins, Richard and Fidler, Sanja and Fouhey, David and Damen, Dima},
  journal=NeurIPS,
  year={2022}
}

@inproceedings{huang2023voxposer,
  title={{VoxPoser: Composable 3D Value Maps for Robotic Manipulation with Language Models}},
  author={Huang, Wenlong and Wang, Chen and Zhang, Ruohan and Li, Yunzhu and Wu, Jiajun and Fei-Fei, Li},
  booktitle={Conference on Robot Learning (CoRL)},
  year={2023}
}

%%%%%%%%%%%%%%%%%%%%%%%%%%%%%%%%%%%%%%%%%%%%%%%%%%%%%%%%%%%%

\appendix
\newpage{}
\section{Benchmark Details}
\label{app:benchmark_detail}

\subsection{Additional Examples}
\label{app:additional_examples}

\begin{figure}[h]
  \centering
  \includegraphics[width=\linewidth]{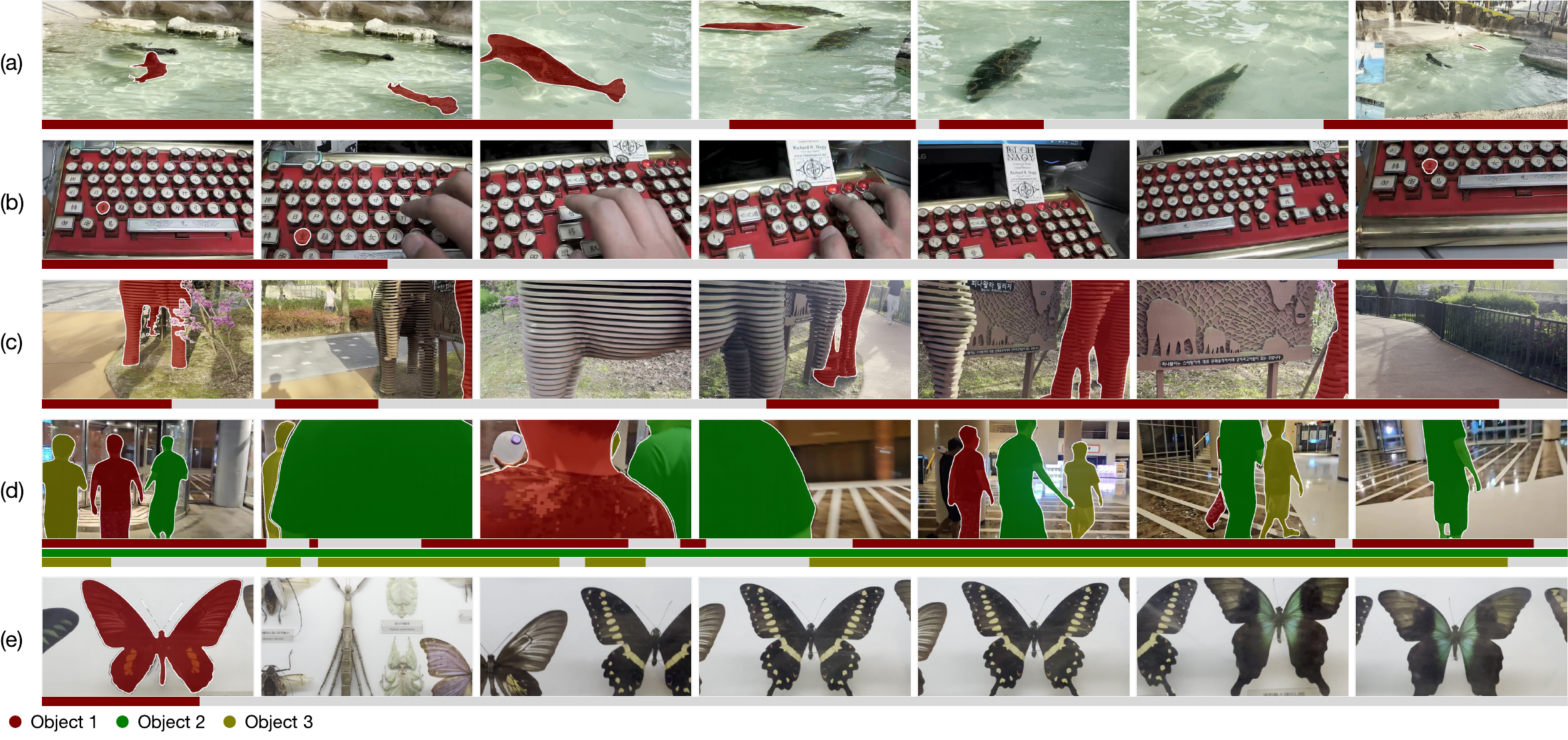}
  \caption{Additional examples from the FaVOS benchmark.}
  \label{fig:app_favos_example}
\end{figure}

Fig.~\ref{fig:app_favos_example} presents additional examples from \benchname{}, illustrating diverse low-temporal-visibility scenarios such as intermittent disappearance, heavy occlusion, large appearance variation, distractor objects, and abrupt motion. The examples highlight that targets are often visible only for a small fraction of the sequence and may reappear after long absence periods, making reliable video object segmentation substantially more challenging than conventional benchmarks with persistently visible objects.

\subsection{Temporal Visibility Distribution}

\begin{figure}[h]
  \centering
  \includegraphics[width=0.6\linewidth]{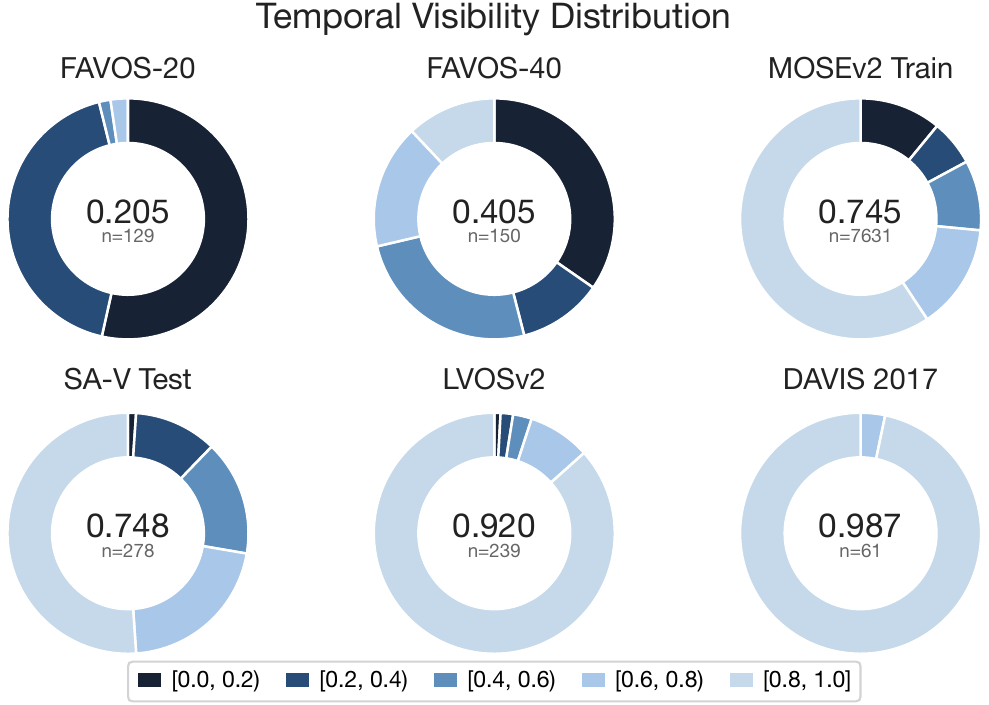}
    \caption{Temporal visibility distributions of different VOS benchmarks. The center value indicates the average temporal visibility, and $n$ denotes the number of annotated objects. Since MOSEv2 test annotations are not publicly available, we report statistics on the training split for MOSEv2. Compared to existing benchmarks, \benchname{} contains substantially more low-visibility objects.}
  \label{fig:app_donut}
\end{figure}

Fig.~\ref{fig:app_donut} compares the temporal visibility distributions of \benchname{} and existing VOS benchmarks. While most existing benchmarks are dominated by highly visible objects, \benchname{} explicitly emphasizes low-temporal-visibility scenarios. In particular, FaVOS-20 and FaVOS-40 are constructed to have average temporal visibility values of approximately 0.2 and 0.4, respectively, resulting in substantially larger portions of objects with frequent disappearance and reappearance.

\subsection{Construction Process}

We construct \benchname{} from two complementary sources: videos filtered from YFCC100M~\cite{thomee2016yfcc100m} and self-collected videos. For the YFCC100M source, we retain valid MP4 videos longer than 20 seconds and filter out nearly static sequences using sparse-frame motion statistics. We further analyze sampled frames with an off-the-shelf object detector to estimate object count, object size, object overlap, and scene complexity. Videos with low resolution, trivial layouts, or a single dominant foreground object are discarded, yielding a candidate pool with richer object interactions, distractors, and non-trivial temporal dynamics. The self-collected videos complement the YFCC100M pool with additional scenarios matching our target low-temporal-visibility setting.

For the remaining videos, we generate initial mask proposals and rank candidate object tracks according to their temporal behavior. We prioritize objects that are visible for a sufficient number of frames, disappear and reappear, exhibit area changes, and are neither persistently visible nor overwhelmingly large. Selected candidates are then manually reviewed and refined with an iterative annotation interface, where annotators provide prompts, propagate masks, and correct masks through additional prompts or direct editing. Unsuitable sequences, including low-quality, harmful, or privacy-sensitive content, are removed, and faces are defaced when necessary. Finally, accepted sequences are converted into the released benchmark format with RGB frames, indexed segmentation masks, and metadata.

\begin{revblock}
\subsection{Split Statistics}
\label{app:split_stats}

\begin{table}[t]
\begin{revblock}
  \centering
  \setlength{\tabcolsep}{9pt}
  \renewcommand{\arraystretch}{1.1}
  \small
  \caption{Statistics of the \benchname{}-20 and \benchname{}-40 splits.}
  \begin{tabular}{l cc}
    \toprule
    \textbf{Statistic} & \textbf{\benchname{}-20} & \textbf{\benchname{}-40} \\
    \midrule
    Videos & 100 & 100 \\
    Objects & 129 & \corr{150} \\
    Objects per video & 1.29 & \corr{1.50} \\
    Videos from YFCC100M / self-collected & 67 / 33 & 66 / 34 \\
    \cmidrule(lr){1-3}
    Mean temporal visibility $\tau$ & 0.21 & \corr{0.40} \\
    Mean disappearance intervals per object & 1.54 & 1.90 \\
    Mean total duration (frames) & 213.0 & 186.0 \\
    Median relative object area & 2.9\% & 2.5\% \\
    \bottomrule
  \end{tabular}
  \label{tab:split_stats}
\end{revblock}
\end{table}

Tab.~\ref{tab:split_stats} summarizes the two splits. The source compositions are nearly identical, the object-size distributions are similar, and the video durations differ only modestly. The mean number of disappearance intervals per object is also comparable, so the two splits differ mainly in absence duration rather than disappearance frequency. \benchname{}-40 contains more objects per video, and thus more object sequences, despite having the same number of videos.
\end{revblock}

\subsection{License and Terms of Use}
\label{app:license_terms}

For videos derived from YFCC100M~\cite{thomee2016yfcc100m}, we retain and follow the original license and terms of use associated with each source video. The extracted frames in the \texttt{JPEGImages} directory are therefore governed by the same licensing conditions as the corresponding YFCC100M videos. Newly collected videos and our newly created segmentation annotations are released under the Creative Commons Attribution-NonCommercial 4.0 International license (CC BY-NC 4.0). The benchmark is intended for research and non-commercial use, and users are responsible for respecting the applicable licenses and terms of use for each asset.

\section{Formula Details}
\label{app:formula_detail}

\subsection{Details of the Contour Accuracy $\mathcal{F}$}
\label{app:f_measure}

Contour accuracy $\mathcal{F}$ evaluates boundary alignment between a predicted mask and its ground-truth mask. Following the DAVIS~\cite{perazzi2016davis} contour evaluation protocol, a binary mask $\mathcal{M}$ is represented by its one-pixel-wide contour set $c(\mathcal{M})$. Let $c(\mathcal{M})$ and $c(\hat{\mathcal{M}})$ denote the ground-truth and predicted contour sets, where two contour points are matched if they fall within a spatial tolerance $\delta_{xy}$.

Let $m(\mathcal{M},\hat{\mathcal{M}})$ denote the number of matched contour points. The contour precision, recall, and contour accuracy are defined as
\begin{equation}
P_c
=
\frac{m(\mathcal{M},\hat{\mathcal{M}})}{|c(\hat{\mathcal{M}})|},
\qquad
R_c
=
\frac{m(\mathcal{M},\hat{\mathcal{M}})}{|c(\mathcal{M})|},
\qquad
\mathcal{F}
=
\frac{2P_cR_c}{P_c+R_c}.
\end{equation}
In the DAVIS implementation, this tolerance-based matching is efficiently approximated by dilating each contour map with a disk structuring element of radius
$\lceil 0.008\sqrt{H^2+W^2}\rceil$ and counting boundary hits within the dilated regions. When both contours are empty, $\mathcal{F}=1$; when only one contour is empty, $\mathcal{F}=0$.

% \subsection{Derivation of \jnf{} Decomposition}
% \label{app:derivation_jnf_decomposition}

% Let $\tau$ denote the average temporal visibility over object-frame pairs in
% the dataset. Thus, a fraction $\tau$ of object-frame pairs contain the target
% object, while the remaining fraction $1-\tau$ correspond to target-absent
% frames.

% On target-present frames, the predictor declares the object present with
% probability $p_1$. When this occurs, it achieves an expected frame-level \jnf{}
% score of $s$; otherwise, it outputs an empty mask and receives zero score.
% Therefore, the expected contribution from target-present frames is
% \begin{equation}
% \tau p_1 s.
% \end{equation}

% On target-absent frames, \jnf{} reduces to a binary empty-mask correctness score:
% the predictor receives score $1$ if it outputs an empty mask, and $0$ otherwise.
% Since this occurs with probability $p_0$, the expected contribution from
% target-absent frames is
% \begin{equation}
% (1-\tau)p_0.
% \end{equation}

% Combining the two cases gives
% \begin{equation}
% \mathjnf
% =
% \tau p_1 s + (1-\tau)p_0.
% \end{equation}

\subsection{Derivation of \jnf{} Decomposition}
\label{app:derivation_jnf_decomposition}

We derive Eq.~\ref{eq:jnf_decomposition} from the standard object-level
averaging used in \jnf{}. Since the dataset-level score is an arithmetic mean
over object sequences, it is sufficient to derive the decomposition for a
single object sequence.

For a single object sequence, let
\begin{equation}
q_t
=
\frac{1}{2}
\left(
\mathcal{J}(\mathcal{M}_t,\hat{\mathcal{M}}_t)
+
\mathcal{F}(\mathcal{M}_t,\hat{\mathcal{M}}_t)
\right)
\end{equation}
denote the frame-level \jnf{} score. The object-level score is then
\begin{equation}
\mathjnf
=
\frac{1}{T}
\sum_{t=1}^{T} q_t .
\end{equation}

Using the ground-truth visibility indicator $e_t$, we decompose the temporal
average into target-present and target-absent frames:
\begin{equation}
\mathjnf
=
\frac{1}{T}
\sum_{t:e_t=1} q_t
+
\frac{1}{T}
\sum_{t:e_t=0} q_t .
\end{equation}

For target-present frames, i.e., $e_t=1$, the prediction contributes to
segmentation quality only when the target is also predicted as present. If
$\hat e_t=0$, the predicted mask is empty and the frame receives zero score.
Thus,
\begin{equation}
q_t
=
\hat e_t
\cdot
\frac{1}{2}
\left(
\mathcal{J}(\mathcal{M}_t,\hat{\mathcal{M}}_t)
+
\mathcal{F}(\mathcal{M}_t,\hat{\mathcal{M}}_t)
\right),
\qquad
\text{for } e_t=1 .
\end{equation}
By the definition of $s$, the average segmentation quality conditioned on
$e_t=1$ and $\hat e_t=1$ is $s$. Therefore,
\begin{equation}
\frac{1}{T}
\sum_{t:e_t=1} q_t
=
\frac{|\mathcal{T}_1|}{T}
\left(
\frac{1}{|\mathcal{T}_1|}
\sum_{t\in\mathcal{T}_1} \hat e_t
\right)
s
=
\tau p_1 s .
\end{equation}

For target-absent frames, i.e., $e_t=0$, the ground-truth mask is empty. Under
the standard \jnf{} convention, both $\mathcal{J}$ and $\mathcal{F}$ reduce to
a binary correctness indicator based on whether the prediction is also empty:
\begin{equation}
q_t
=
\mathbf{1}\left(\|\hat{\mathcal{M}}_t\|_0=0\right)
=
1-\hat e_t,
\qquad
\text{for } e_t=0 .
\end{equation}
Hence, the contribution from target-absent frames is
\begin{equation}
\frac{1}{T}
\sum_{t:e_t=0} q_t
=
\frac{|\mathcal{T}_0|}{T}
\left(
\frac{1}{|\mathcal{T}_0|}
\sum_{t\in\mathcal{T}_0}
(1-\hat e_t)
\right)
=
(1-\tau)p_0 .
\end{equation}

Combining the target-present and target-absent contributions gives
\begin{equation}
\mathjnf
=
\tau p_1 s
+
(1-\tau)p_0 .
\end{equation}
Thus, the standard \jnf{} score decomposes into a visible-frame
segmentation term $\tau p_1 s$ and a target-absence correctness term
$(1-\tau)p_0$.

\subsection{Alternative Metrics of \jnf{}}
\label{app:alternative_jnf}

The standard evaluation metric for VOS is \jnf{}, which combines region similarity $\mathcal{J}$, based on intersection-over-union (IoU)~[\citenum{everingham2010pascalvoc}, \rev{\citenum{ravi2024sam2}, \citenum{kim2026dpseg}, \citenum{carion2025sam3}}], and contour accuracy $\mathcal{F}$, following the DAVIS evaluation protocol~\cite{perazzi2016davis}. This metric has been widely adopted across VOS variants, including semi-supervised VOS~\cite{caelles2017osvos,perazzi2016davis}, unsupervised VOS~\cite{caelles2019uvos}, referring VOS~\cite{khoreva2018refdavis,gavrilyuk2018a2dsentences,ding2023mevis}, and audio-guided VOS~\cite{zhou2022avos,pan2022wnet,ding2025mevisv2}. However, \jnf{} is computed independently on each frame and then averaged over time, limiting its ability to capture temporal structure such as disappearance, reappearance, and prediction stability.

Several alternative metrics have therefore been proposed to incorporate temporal behavior and are commonly reported alongside \jnf{} as supplementary diagnostics. In this section, we describe the definitions and characteristics of these metrics.

\paragraph{Temporal stability. }
Temporal stability $\mathcal{T}$~\cite{perazzi2016davis} evaluates the temporal coherence of predicted object contours by penalizing jittery and unstable boundary variations across consecutive frames.

Given a predicted mask $M_t$ at frame $t$, the contour is first represented as a polygon:
\begin{equation}
P(M_t)=\{p_t^i\}_{i=1}^{N_t},
\end{equation}
where each contour point $p_t^i$ is described using a Shape Context Descriptor (SCD)~\cite{shape2002scd}. 

To compensate for object motion and smooth deformations, contour matching between consecutive frames is formulated as a Dynamic Time Warping (DTW)~\cite{speech1993dtw} problem:
\begin{equation}
\pi^\ast
=
\arg\min_{\pi}
\sum_{(i,j)\in\pi}
d_{\mathrm{SCD}}(p_t^i,p_{t+1}^j),
\end{equation}
where $\pi$ denotes an order-preserving matching path and $d_{\mathrm{SCD}}(\cdot,\cdot)$ represents the distance between shape context descriptors.

The temporal stability measure is then defined as the average matching cost:
\begin{equation}
\mathcal{T}
=
\frac{1}{|\pi^\ast|}
\sum_{(i,j)\in\pi^\ast}
d_{\mathrm{SCD}}(p_t^i,p_{t+1}^j).
\end{equation}

Intuitively, the matching compensates for smooth object motion and moderate deformations, while unstable contour oscillations and boundary inaccuracies increase the matching cost. Lower $\mathcal{T}$ therefore indicates more temporally stable predictions.

Unlike frame-wise segmentation metrics, $\mathcal{T}$ explicitly captures temporal coherence across frames. However, it mainly evaluates contour smoothness and does not directly measure object presence consistency under disappearance and reappearance scenarios.

\paragraph{$\boldsymbol{\mathcal{J}\&\dot{\mathcal{F}}}$. }
The standard contour metric $\mathcal{F}$ in DAVIS uses a boundary tolerance proportional to the image diagonal:
\begin{equation}
\omega = 0.008 \times \mathcal{D},
\end{equation}
where $\mathcal{D}$ denotes the image diagonal length. 
This formulation tends to favor large objects since the tolerance region grows with image size regardless of object scale.

To alleviate this bias, MOSEv2~\cite{mosev2} introduces a modified contour metric $\dot{\mathcal{F}}$ using an adaptive tolerance:
\begin{equation}
\dot{\omega} = \min(0.008 \times \mathcal{D}, \; 0.1 \times \sqrt{A}),
\end{equation}
where $A$ denotes the object area. 
The resulting metric is defined as
\begin{equation}
\mathcal{J}\&\dot{\mathcal{F}}
=
\frac{1}{2}
\left(
\mathcal{J}
+
\dot{\mathcal{F}}
\right).
\end{equation}
By constraining the boundary tolerance according to object size, $\dot{\mathcal{F}}$ reduces the overestimation of contour quality on small objects.

\paragraph{$\boldsymbol{\mathcal{J}\&\dot{\mathcal{F}}_d}$ and
$\boldsymbol{\mathcal{J}\&\dot{\mathcal{F}}_r}$.}
MOSEv2~\cite{mosev2} further introduces metrics specialized for disappearance
and reappearance clips. Let
\begin{equation}
\mathcal{C}_d = \{I_d^{(m)}\}_{m=1}^{N_d},
\qquad
\mathcal{C}_r = \{I_r^{(m)}\}_{m=1}^{N_r}
\end{equation}
denote the sets of disappearance and reappearance clips, respectively. Here,
each $I_d^{(m)}$ is a contiguous interval where the target is absent, and each
$I_r^{(m)}$ is a contiguous target-present interval after a disappearance. The
initial continuous target-present segment is excluded from $\mathcal{C}_r$.

The interval-aware metrics are computed by first averaging within each clip and
then averaging over clips:
\begin{equation}
\mathcal{J}\&\dot{\mathcal{F}}_d
=
\frac{1}{N_d}
\sum_{m=1}^{N_d}
\frac{1}{|I_d^{(m)}|}
\sum_{t \in I_d^{(m)}}
\frac{
\mathcal{J}_t + \dot{\mathcal{F}}_t
}{2},
\end{equation}
and
\begin{equation}
\mathcal{J}\&\dot{\mathcal{F}}_r
=
\frac{1}{N_r}
\sum_{m=1}^{N_r}
\frac{1}{|I_r^{(m)}|}
\sum_{t \in I_r^{(m)}}
\frac{
\mathcal{J}_t + \dot{\mathcal{F}}_t
}{2}.
\end{equation}

These metrics explicitly evaluate model behavior during disappearance and
reappearance clips. However, they remain auxiliary to the standard \jnf{}
protocol, which still exhibits an imbalance between $s$ and $p_0$
(Sec.~\ref{sec.5.1}). Moreover, because the final score averages over clips,
short and long disappearance or reappearance events can receive similar
sequence-level weight.

\paragraph{tIoU and vIoU.}
Long-RVOS~\cite{liang2026longrvos} introduces tIoU and vIoU to explicitly
evaluate temporal presence overlap and mask overlap over temporally aligned
object-support frames.

For predicted masks $\hat{\mathcal{M}}_t$ and ground-truth masks
$\mathcal{M}_t$, the predicted and ground-truth temporal support sets are
defined as
\begin{equation}
    \hat{\mathcal{S}}
    =
    \{t \mid \|\hat{\mathcal{M}}_t\|_0 > 0\},
    \qquad
    \mathcal{S}
    =
    \{t \mid \|\mathcal{M}_t\|_0 > 0\},
\end{equation}
where $\|\cdot\|_0$ denotes the number of non-zero pixels.

Temporal IoU measures the overlap between predicted and ground-truth temporal
support:
\begin{equation}
    \mathrm{tIoU}
    =
    \frac{
        |\hat{\mathcal{S}} \cap \mathcal{S}|
    }{
        |\hat{\mathcal{S}} \cup \mathcal{S}|
    }.
\end{equation}

The vIoU metric evaluates mask overlap over frames where both predicted and
ground-truth temporal supports overlap:
\begin{equation}
    \mathrm{vIoU}
    =
    \frac{1}{
        |\hat{\mathcal{S}} \cup \mathcal{S}|
    }
    \sum_{t \in \hat{\mathcal{S}} \cap \mathcal{S}}
    \mathcal{J}(\mathcal{M}_t,\hat{\mathcal{M}}_t).
\end{equation}

These metrics explicitly evaluate temporal presence overlap and mask quality
within temporally overlapping supports. However, they do not account for contour
accuracy, and are therefore used as supplementary metrics alongside the standard
\jnf{} protocol.

\paragraph{Volumetric IoU.}
Related volumetric measures are also used in video instance segmentation
(VIS)~\cite{yang2019vis}, where predictions are commonly evaluated using
volumetric IoU and average precision.

Given ground-truth and predicted mask sequences
$\mathcal{M}=\{\mathcal{M}_t\}_{t=1}^{T}$ and
$\hat{\mathcal{M}}=\{\hat{\mathcal{M}}_t\}_{t=1}^{T}$, let
$\mathcal{V}$ and $\hat{\mathcal{V}}$ denote the corresponding
spatio-temporal mask volumes. The volumetric IoU is computed over the entire
spatio-temporal volume as
\begin{equation}
    \mathrm{VIS\text{-}IoU}(\mathcal{V},\hat{\mathcal{V}})
    =
    \frac{
        \|\mathcal{V} \cap \hat{\mathcal{V}}\|_0
    }{
        \|\mathcal{V} \cup \hat{\mathcal{V}}\|_0
    }
    =
    \frac{
        \sum_{t=1}^{T}
        \|\mathcal{M}_t \cap \hat{\mathcal{M}}_t\|_0
    }{
        \sum_{t=1}^{T}
        \|\mathcal{M}_t \cup \hat{\mathcal{M}}_t\|_0
    }.
\end{equation}

While VIS-IoU naturally handles spatio-temporal mask tubes, it is typically
used within detection-style evaluation with confidence-ranked predictions and
average precision. In contrast, \jnfv{} preserves the region-and-contour
structure of the standard \jnf{} protocol.

\subsection{Additional \jnf{} Contribution Analysis}
\label{app:jnf_contribution}

\begin{figure}[!t]
  \centering
  \includegraphics[width=\linewidth]{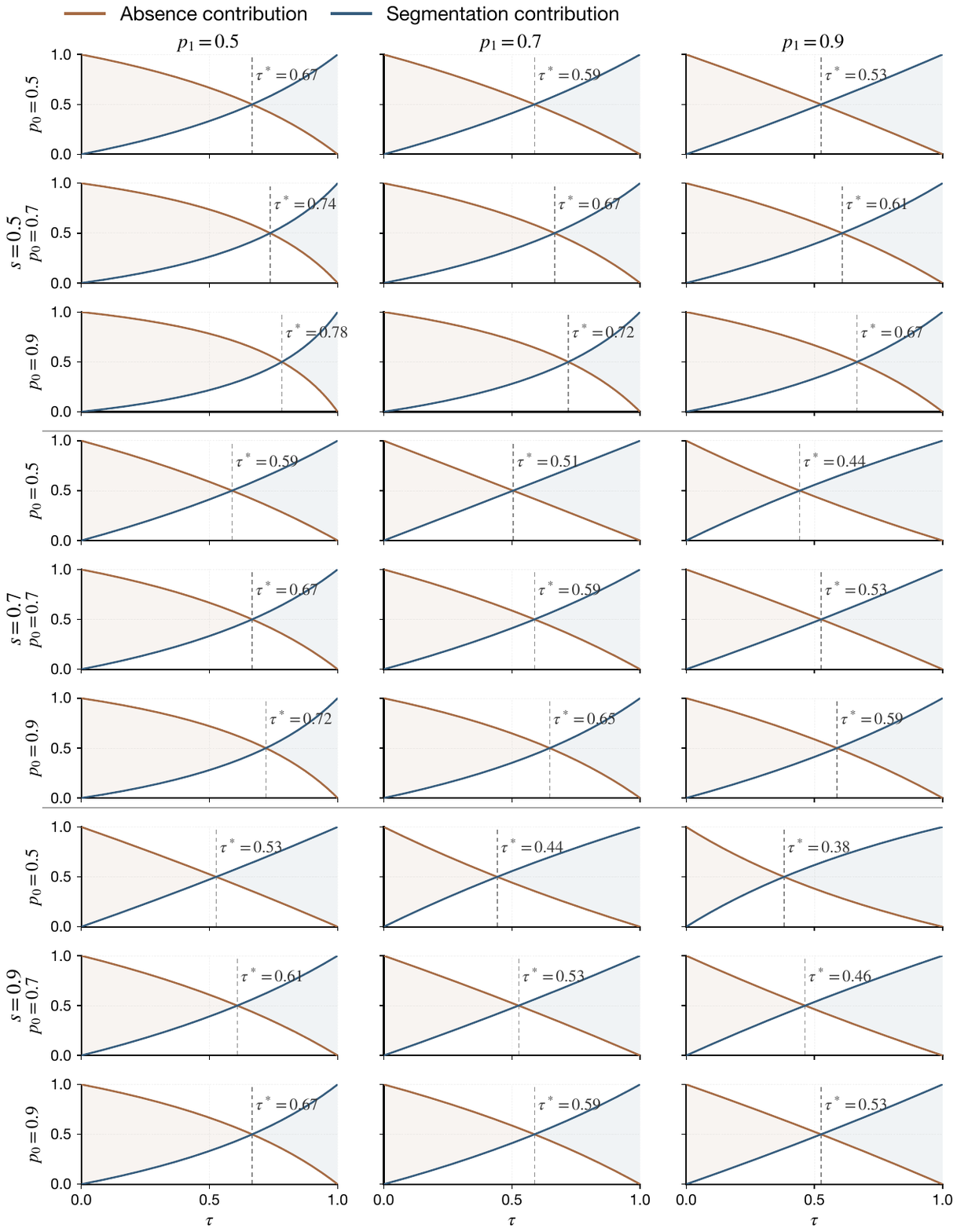}
    \caption{Relative contributions of the absence and segmentation terms in \jnf{} across different $p_0$, $p_1$, and $s$ settings. The dashed line marks the crossover visibility $\tau^*$ where the two contributions are equal.}
  \label{fig:app_jnf_contribution}
\end{figure}

Figure~\ref{fig:app_jnf_contribution} shows the contribution ratios of the absence and segmentation terms under different $p_0$, $p_1$, and $s$ settings. Across all settings, decreasing temporal visibility shifts the dominant contribution from segmentation quality to absence classification. Higher $p_0$ expands the absence-dominated regime, while higher $p_1$ or $s$ shifts dominance toward the segmentation term. Nevertheless, once temporal visibility falls below a moderate range, the absence term often becomes dominant, indicating that low-visibility benchmarks require evaluation metrics that explicitly account for this imbalance.

\subsection{Derivation of Volumetric Jaccard}
\label{app:derivation_volumetric_jaccard}

Given a ground-truth mask sequence $\mathcal{M}=\{\mathcal{M}_t\}_{t=1}^{T}$ and a predicted mask sequence $\hat{\mathcal{M}}=\{\hat{\mathcal{M}}_t\}_{t=1}^{T}$, we define their spatio-temporal mask volumes by stacking the 2D masks along the temporal axis:
\begin{equation}
    \mathcal{V}(x,y,t)=\mathcal{M}_t(x,y),
    \qquad
    \hat{\mathcal{V}}(x,y,t)=\hat{\mathcal{M}}_t(x,y).
\end{equation}
The volumetric Jaccard index is defined as the IoU between these two binary volumes:
\begin{equation}
    \mathcal{J}_{v}
    =
    \frac{\|\mathcal{V}\cap\hat{\mathcal{V}}\|_0}
         {\|\mathcal{V}\cup\hat{\mathcal{V}}\|_0}.
\end{equation}

Since the volumes are constructed by stacking frame-wise masks, their intersection and union decompose over time:
\begin{equation}
    \|\mathcal{V}\cap\hat{\mathcal{V}}\|_0
    =
    \sum_{t=1}^{T}
    \|\mathcal{M}_t\cap\hat{\mathcal{M}}_t\|_0,
\end{equation}
\begin{equation}
    \|\mathcal{V}\cup\hat{\mathcal{V}}\|_0
    =
    \sum_{t=1}^{T}
    \|\mathcal{M}_t\cup\hat{\mathcal{M}}_t\|_0.
\end{equation}
Therefore,
\begin{equation}
    \mathcal{J}_{v}
    =
    \frac{
        \sum_{t=1}^{T}
        \|\mathcal{M}_t\cap\hat{\mathcal{M}}_t\|_0
    }{
        \sum_{t=1}^{T}
        \|\mathcal{M}_t\cup\hat{\mathcal{M}}_t\|_0
    }.
\end{equation}

Let
\begin{equation}
    u_t = \|\mathcal{M}_t\cup\hat{\mathcal{M}}_t\|_0,
\end{equation}
and define the per-frame Jaccard index as
\begin{equation}
    \mathcal{J}_t
    =
    \mathcal{J}(\mathcal{M}_t,\hat{\mathcal{M}}_t)
    =
    \frac{\|\mathcal{M}_t\cap\hat{\mathcal{M}}_t\|_0}
         {\|\mathcal{M}_t\cup\hat{\mathcal{M}}_t\|_0}.
\end{equation}
For frames with $u_t=0$, both masks are empty and the frame contributes neither to the volumetric intersection nor to the volumetric union. Thus, such frames can be ignored in the weighted form. For frames with $u_t>0$, we have
\begin{equation}
    \|\mathcal{M}_t\cap\hat{\mathcal{M}}_t\|_0
    =
    u_t \mathcal{J}_t.
\end{equation}
Substituting this into the volumetric IoU gives
\begin{equation}
    \mathcal{J}_{v}
    =
    \frac{
        \sum_{t=1}^{T}
        u_t \mathcal{J}_t
    }{
        \sum_{t=1}^{T}
        u_t
    }.
\end{equation}
Hence, the volumetric Jaccard index is exactly a union-volume-weighted average of frame-wise Jaccard scores.

More generally, a weighted temporal aggregation of frame-wise Jaccard scores can be written as
\begin{equation}
    \mathcal{J}_{w}
    =
    \frac{
        \sum_{t=1}^{T}
        w(u_t)\mathcal{J}_t
    }{
        \sum_{t=1}^{T}
        w(u_t)
    }.
\end{equation}
Choosing $w(u_t)=u_t$ recovers the volumetric Jaccard index $\mathcal{J}_{v}$, which corresponds to a VIS-style spatio-temporal IoU over mask tubes~\cite{yang2019vis}. In contrast, choosing a constant weight over temporal-union frames, e.g., $w(u_t)=\mathbf{1}(u_t>0)$, yields an unweighted temporal average of frame-wise IoU over frames where either the prediction or ground truth is non-empty, matching the vIoU-style aggregation used in long-video VOS~\cite{liang2026longrvos}.

\subsection{Implementation of Volumetric Contour Accuracy}
\label{app:volumetric_contour_accuracy}

The volumetric contour accuracy $\mathcal{F}_v$ follows the DAVIS-style contour matching definition, where boundary precision and recall are defined through bipartite matching between ground-truth and predicted boundary elements. Following the DAVIS implementation, we approximate this matching using morphology-equivalent tolerance regions.

Given ground-truth and predicted mask volumes $\mathcal{V},\hat{\mathcal{V}}\in\{0,1\}^{T\times H\times W}$, we first extract boundary voxel sets $c(\mathcal{V})$ and $c(\hat{\mathcal{V}})$ by detecting label changes between neighboring voxels in the spatio-temporal volume. Thus, the extracted boundaries include both spatial object contours and temporal changes in object support. Boundary matching is then approximated using a tolerance operator $\mathcal{D}_{\delta_{xy},\delta_t}(\cdot)$, where $\delta_{xy}$ controls the spatial matching tolerance and $\delta_t$ controls the temporal matching tolerance. In our default evaluation, we use same-frame boundary matching, corresponding to $\delta_t=0$. We also observe that using $\delta_t>0$ preserves the same qualitative trends while yielding more permissive scores.

For the spatial tolerance, we follow the standard boundary-tolerance setting and cap it by object scale following MOSEv2~\cite{mosev2}:
\begin{equation}
\delta_{xy}
=
\left\lceil
\min\left(
\omega\sqrt{H^2+W^2},
\alpha\sqrt{A}
\right)
\right\rceil,
\end{equation}
where $\omega$ is the image-diagonal tolerance ratio, $A$ is the mean visible object area, and $\alpha$ controls the object-scale cap. We use $\omega=0.008$ and $\alpha=0.1$.

Using the morphology-equivalent tolerance operator, the matched predicted and ground-truth boundary voxels are approximated as
\begin{equation}
m_{\mathrm{pred}}
=
\left|
c(\hat{\mathcal{V}})
\cap
\mathcal{D}_{\delta_{xy},\delta_t}
\left(c(\mathcal{V})\right)
\right|,
\qquad
m_{\mathrm{gt}}
=
\left|
c(\mathcal{V})
\cap
\mathcal{D}_{\delta_{xy},\delta_t}
\left(c(\hat{\mathcal{V}})\right)
\right|.
\end{equation}

The resulting volumetric contour precision and recall are
\begin{equation}
P_c^{v}
=
\frac{m_{\mathrm{pred}}}{|c(\hat{\mathcal{V}})|},
\qquad
R_c^{v}
=
\frac{m_{\mathrm{gt}}}{|c(\mathcal{V})|}.
\end{equation}
Finally, the volumetric contour accuracy is computed as
\begin{equation}
\mathcal{F}_{v}
=
\frac{2P_c^{v}R_c^{v}}{P_c^{v}+R_c^{v}}.
\end{equation}

For empty boundary cases, we follow the standard convention: if both boundary sets are empty, we set $P_c^{v}=R_c^{v}=1$; if only the predicted boundary set is empty, we set $P_c^{v}=1$ and $R_c^{v}=0$; and if only the ground-truth boundary set is empty, we set $P_c^{v}=0$ and $R_c^{v}=1$.

\section{Experimental Details}
\label{app:experimental_details}
All reported metric evaluations were run on a local research compute server with two AMD EPYC 7763 64-core CPUs, 1\,TiB system memory, and 8 NVIDIA A100 GPUs with 80\,GB memory each. Metric computation is CPU-based and does not require GPU memory; GPUs are only used to generate model predictions before evaluation.

\rev{Following the semi-supervised VOS setting, all models receive the same first-frame ground-truth mask as the prompt.} For \jnfv{} and related video-object metrics, we evaluate each video-object sequence independently and parallelize across sequences using Python multiprocessing with 16 CPU workers. Each worker loads the corresponding ground-truth and predicted PNG mask sequences and computes object-level $\mathcal{J}_v$, $\mathcal{F}_v$, and \jnfv{} using the reported settings, including the DAVIS-style boundary tolerance of $0.008$. The memory footprint is mainly determined by the binary mask volumes loaded for the current sequences, and all evaluations fit within the available host memory.

\begin{revblock}
\section{Additional Results}
\label{app:additional_results}

\subsection{Backward Compatibility on High-Visibility Benchmarks}
\label{app:backward_compat}

\begin{table}[t]
\begin{revblock}
  \centering
  \setlength{\tabcolsep}{4pt}
  \renewcommand{\arraystretch}{0.95}
  \footnotesize
  \caption{Comparison of \jnf{} and \jnfv{} on high-temporal-visibility benchmarks. $\Delta$ denotes \jnfv{} $-$ \jnf{}. Rows are sorted by \jnf{} on DAVIS17.}
  {\setlength{\arrayrulewidth}{0.3pt}
    \begin{tabular}{l|c c c|c c c}
      \toprule
      \textbf{Model}
      & \multicolumn{3}{c|}{\textbf{DAVIS17 val}}
      & \multicolumn{3}{c}{\textbf{OVOS val}} \\
      \cmidrule(lr){2-4} \cmidrule(lr){5-7}
      & \jnf{}$\uparrow$ & \jnfv{}$\uparrow$ & $\Delta$
      & \jnf{}$\uparrow$ & \jnfv{}$\uparrow$ & $\Delta$ \\
      \midrule

      SAM 3~\cite{carion2025sam3}
        & 91.47 & 93.15 & +1.68
        & 86.67 & 87.00 & +0.33 \\

      SeC~\cite{zhang2025secvos}
        & 89.87 & 91.28 & +1.42
        & 81.62 & 81.81 & +0.19 \\

      SAM 2.1-L~\cite{ravi2024sam2}
        & 89.80 & 91.38 & +1.58
        & 80.98 & 81.22 & +0.24 \\

      SAM2Long-L~\cite{ding2025sam2long}
        & 89.80 & 91.35 & +1.55
        & 81.45 & 81.77 & +0.32 \\

      SAMURAI-L~\cite{yang2024samurai}
        & 89.57 & 91.19 & +1.62
        & 79.68 & 80.29 & +0.61 \\

      DAM4SAM-L~\cite{videnovic2025didi}
        & 88.64 & 90.15 & +1.51
        & 80.29 & 80.72 & +0.43 \\

      Cutie-B~\cite{cheng2024cutie}
        & 88.04 & 89.17 & +1.13
        & 80.77 & 80.96 & +0.19 \\

      XMem~\cite{cheng2022xmem}
        & 86.15 & 87.49 & +1.34
        & 67.63 & 67.38 & $-$0.25 \\

      STCN~\cite{cheng2021stcn}
        & 85.29 & 86.40 & +1.11
        & 61.27 & 60.57 & $-$0.70 \\

      DeAOT-L~\cite{yang2022deaot}
        & 84.01 & 85.28 & +1.27
        & 68.25 & 68.49 & +0.24 \\

      STM~\cite{oh2019stm}
        & 81.70 & 83.94 & +2.24
        & 57.04 & 56.57 & $-$0.47 \\

      \rowcolor{blue!8}
      \textbf{Empt. pred.}
        & 1.28 & 0.00 & $-$1.28
        & 11.13 & 0.07 & $-$11.06 \\

      \bottomrule
    \end{tabular}
  }
  \label{tab:backward_compat}
\end{revblock}
\end{table}

To examine whether \jnfv{} is backward compatible with standard \jnf{} evaluation in conventional high-visibility settings, we additionally evaluate all models on the DAVIS17 validation set~\cite{perazzi2016davis} (30 videos and 61 objects) and the OVOS validation set~\cite{miao2024ovos} (607 videos and 2,694 objects), whose average temporal visibility is 0.99 and 0.89, respectively. As shown in Tab.~\ref{tab:backward_compat}, \jnfv{} closely follows \jnf{} on both benchmarks. On DAVIS17, the maximum rank change between the two metrics is two positions, and on OVOS the model rankings are preserved exactly. SAM~3 remains the top-performing model on both benchmarks under either metric, while the empty-mask predictor drops to near zero under \jnfv{}.

\subsection{Comparison with Alternative Evaluation Protocols}
\label{app:alternative_protocols}

\begin{table}[t]
\begin{revblock}
  \centering
  \setlength{\tabcolsep}{3.4pt}
  \renewcommand{\arraystretch}{1.08}
  \footnotesize
  \caption{Comparison with alternative evaluation protocols on FaVOS-full and SA-V~\cite{ravi2024sam2} test split. Single-score protocols: standard \jnf{}, visibility-balanced \jnf{} (Bal.), and \jnfv{}, with ranks in parentheses. Two-number protocol: \jnf{} on target-present frames only (Present) and frame-level presence F1. Higher is better for all columns. Rows are sorted by \jnfv{} on FaVOS-full.}
  \begin{tabular}{l|ccc|cc|ccc|cc}
    \toprule
    & \multicolumn{5}{c|}{\textbf{FaVOS-full}} & \multicolumn{5}{c}{\textbf{SA-V test}} \\
    \cmidrule(lr){2-6} \cmidrule(lr){7-11}
    & \multicolumn{3}{c|}{Single score} & \multicolumn{2}{c|}{Two-number} & \multicolumn{3}{c|}{Single score} & \multicolumn{2}{c}{Two-number} \\
    \cmidrule(lr){2-4} \cmidrule(lr){5-6} \cmidrule(lr){7-9} \cmidrule(lr){10-11}
    \textbf{Model} & \jnf{} & Bal. & \jnfv{} & Present & F1 & \jnf{} & Bal. & \jnfv{} & Present & F1 \\
    \midrule
    SAM 3 & 81.8\,{\scriptsize(2)} & 81.3\,{\scriptsize(1)} & 65.0\,{\scriptsize(1)} & 78.4 & 74.5 & 84.7\,{\scriptsize(1)} & 82.9\,{\scriptsize(1)} & 82.6\,{\scriptsize(1)} & 81.5 & 94.0 \\
    SeC & 82.4\,{\scriptsize(1)} & 81.2\,{\scriptsize(2)} & 63.9\,{\scriptsize(2)} & 74.6 & 72.7 & 81.7\,{\scriptsize(2)} & 81.3\,{\scriptsize(2)} & 77.8\,{\scriptsize(2)} & 76.1 & 91.2 \\
    SAM2Long-L & 76.3\,{\scriptsize(5)} & 76.2\,{\scriptsize(4)} & 57.9\,{\scriptsize(3)} & 73.1 & 70.3 & 80.4\,{\scriptsize(3)} & 79.2\,{\scriptsize(4)} & 76.4\,{\scriptsize(4)} & 75.7 & 91.3 \\
    DAM4SAM-L & 74.1\,{\scriptsize(6)} & 75.8\,{\scriptsize(5)} & 54.5\,{\scriptsize(4)} & 75.4 & 66.8 & 80.1\,{\scriptsize(4)} & 79.6\,{\scriptsize(3)} & 76.6\,{\scriptsize(3)} & 74.9 & 90.5 \\
    Cutie-B & 78.4\,{\scriptsize(3)} & 74.4\,{\scriptsize(6)} & 54.2\,{\scriptsize(5)} & 57.9 & 67.9 & 62.2\,{\scriptsize(7)} & 67.7\,{\scriptsize(7)} & 53.4\,{\scriptsize(9)} & 50.7 & 71.7 \\
    SAM 2.1-L & 77.0\,{\scriptsize(4)} & 76.9\,{\scriptsize(3)} & 54.1\,{\scriptsize(6)} & 70.4 & 66.5 & 79.7\,{\scriptsize(6)} & 78.7\,{\scriptsize(5)} & 74.3\,{\scriptsize(6)} & 74.1 & 89.4 \\
    SAMURAI-L & 66.9\,{\scriptsize(9)} & 71.3\,{\scriptsize(7)} & 46.7\,{\scriptsize(7)} & 79.0 & 60.9 & 79.9\,{\scriptsize(5)} & 74.8\,{\scriptsize(6)} & 76.4\,{\scriptsize(5)} & 78.7 & 93.1 \\
    DeAOT-L & 61.4\,{\scriptsize(11)} & 62.2\,{\scriptsize(9)} & 44.0\,{\scriptsize(8)} & 61.6 & 55.6 & 62.2\,{\scriptsize(8)} & 60.1\,{\scriptsize(10)} & 55.9\,{\scriptsize(7)} & 56.9 & 81.7 \\
    XMem & 62.8\,{\scriptsize(10)} & 61.3\,{\scriptsize(10)} & 41.4\,{\scriptsize(9)} & 52.5 & 57.7 & 58.0\,{\scriptsize(10)} & 63.1\,{\scriptsize(8)} & 48.9\,{\scriptsize(10)} & 46.4 & 71.5 \\
    STM & 68.6\,{\scriptsize(8)} & 62.9\,{\scriptsize(8)} & 40.5\,{\scriptsize(10)} & 42.5 & 56.0 & 46.6\,{\scriptsize(11)} & 56.6\,{\scriptsize(11)} & 34.4\,{\scriptsize(11)} & 30.7 & 58.3 \\
    STCN & 52.4\,{\scriptsize(12)} & 53.1\,{\scriptsize(11)} & 32.3\,{\scriptsize(11)} & 51.8 & 52.1 & 60.4\,{\scriptsize(9)} & 62.1\,{\scriptsize(9)} & 55.1\,{\scriptsize(8)} & 55.1 & 77.7 \\
    \rowcolor{blue!8}
    \textbf{Empt. pred.} & 69.2\,{\scriptsize(7)} & 50.0\,{\scriptsize(12)} & 0.0\,{\scriptsize(12)} & 0.0 & 0.0 & 25.3\,{\scriptsize(12)} & 50.0\,{\scriptsize(12)} & 0.0\,{\scriptsize(12)} & 0.0 & 0.0 \\
    \bottomrule
  \end{tabular}
  \label{tab:alternative_protocols}
\end{revblock}
\end{table}

We further compare \jnfv{} with two alternative evaluation protocols in Tab.~\ref{tab:alternative_protocols}. The first is a visibility-balanced \jnf{}, which averages frame-wise \jnf{} separately over target-present and target-absent frames and weights the two equally. The second is a two-number protocol that reports present-frame \jnf{}, computed only over target-present frames, together with presence F1, which measures how accurately each model predicts whether the target is present in each frame.

Visibility-balanced \jnf{} changes the rankings substantially, but the empty-mask predictor still receives 50.0 because perfect absence prediction retains half of the score. It is also defined only for sequences containing both target-present and target-absent frames, covering 277 of \corr{279} \benchname{} objects and 235 of 278 SA-V objects. The two-number protocol provides useful diagnostic information but does not define a unique ordering. On \benchname{}-full, SAMURAI-L obtains the highest present-frame \jnf{} (79.0) but a much lower presence F1 (60.9), whereas SAM~3 achieves 78.4 and 74.5, respectively. In contrast, \jnfv{} is defined for all sequences, assigns zero to the empty-mask predictor, penalizes false positives on target-absent frames by their spatial volume, and provides a single model ranking.

\end{revblock}

\section{Societal Impacts}
\label{app:societal_impact}

This work aims to improve the evaluation of video object segmentation under low temporal visibility, which may benefit applications such as robotics, assistive systems, video editing, and scientific video analysis. However, VOS can also be used in privacy-sensitive settings such as surveillance or large-scale video analytics. To reduce potential risks in our released benchmark, we apply privacy-preserving and content-filtering procedures, including face defacing and removal of potentially harmful, inappropriate, or low-quality videos. We do not release a trained deployment model, and we encourage responsible use of the benchmark and metrics under appropriate privacy, safety, and legal constraints.

%%%%%%%%%%%%%%%%%%%%%%%%%%%%%%%%%%%%%%%%%%%%%%%%%%%%%%%%%%%%

% NeurIPS paper checklist: omitted from the arXiv version; set \showchecklisttrue to include it.
\newif\ifshowchecklist
\showchecklistfalse
\ifshowchecklist
  \newpage
  \section*{NeurIPS Paper Checklist}

\begin{enumerate}

\item {\bf Claims}
    \item[] Question: Do the main claims made in the abstract and introduction accurately reflect the paper's contributions and scope?
    \item[] Answer: \answerYes{} % Replace by \answerYes{}, \answerNo{}, or \answerNA{}.
    \item[] Justification: The paper's main claims are that standard \jnf{} collapses under low temporal visibility and that \jnfv{} mitigates this issue. We support the first claim both theoretically (Section~\ref{sec.5.1}) and empirically (Section~\ref{sec:experiments}). We further validate the effectiveness of \jnfv{} through empirical comparisons and analysis in Section~\ref{sec:experiments}.

\item {\bf Limitations}
    \item[] Question: Does the paper discuss the limitations of the work performed by the authors?
    \item[] Answer: \answerYes{} % Replace by \answerYes{}, \answerNo{}, or \answerNA{}.
    \item[] Justification: \rev{We discuss the limitations of our work in Section~\ref{sec:limitations}.}

\item {\bf Theory assumptions and proofs}
    \item[] Question: For each theoretical result, does the paper provide the full set of assumptions and a complete (and correct) proof?
    \item[] Answer: \answerYes{} % Replace by \answerYes{}, \answerNo{}, or \answerNA{}.
    \item[] Justification: We present the decomposition of the standard \jnf{} score in Eq.~\ref{eq:jnf_decomposition} and provide its full derivation in Appendix~\ref{app:derivation_jnf_decomposition}. The \jnf{} collapse contribution in Sec.~\ref{sec.5.1} is then computed from this decomposition, with additional settings and analyses provided in Appendix~\ref{app:jnf_contribution}.

    \item {\bf Experimental result reproducibility}
    \item[] Question: Does the paper fully disclose all the information needed to reproduce the main experimental results of the paper to the extent that it affects the main claims and/or conclusions of the paper (regardless of whether the code and data are provided or not)?
    \item[] Answer: \answerYes{} % Replace by \answerYes{}, \answerNo{}, or \answerNA{}.
    \item[] Justification: We provide the code and datasets as supplementary materials, including the information needed to reproduce the main experimental results supporting our claims.

\item {\bf Open access to data and code}
    \item[] Question: Does the paper provide open access to the data and code, with sufficient instructions to faithfully reproduce the main experimental results, as described in supplemental material?
    \item[] Answer: \answerYes{} % Replace by \answerYes{}, \answerNo{}, or \answerNA{}.
    \item[] Justification: We provide the code through GitHub and release the dataset, including examples and instructions, through Hugging Face as described in the supplemental material.

\item {\bf Experimental setting/details}
    \item[] Question: Does the paper specify all the training and test details (e.g., data splits, hyperparameters, how they were chosen, type of optimizer) necessary to understand the results?
    \item[] Answer: \answerYes{} % Replace by \answerYes{}, \answerNo{}, or \answerNA{}.
    \item[] Justification: Our experiments do not involve additional training. For evaluation, we specify the tested models, datasets, data splits, metrics, and evaluation settings needed to understand and reproduce the reported results.

\item {\bf Experiment statistical significance}
    \item[] Question: Does the paper report error bars suitably and correctly defined or other appropriate information about the statistical significance of the experiments?
    \item[] Answer: \answerNo{} % Replace by \answerYes{}, \answerNo{}, or \answerNA{}.
    \item[] Justification: We report results over the full evaluation splits but do not include error bars or statistical significance tests. Since our experiments evaluate fixed pretrained models without additional training, seed-based variance is not applicable.

\item {\bf Experiments compute resources}
    \item[] Question: For each experiment, does the paper provide sufficient information on the computer resources (type of compute workers, memory, time of execution) needed to reproduce the experiments?
    \item[] Answer: \answerYes{} % Replace by \answerYes{}, \answerNo{}, or \answerNA{}.
    \item[] Justification: We provide the compute resources used for the experiments, including the number of workers, memory requirements, and evaluation settings, in Appendix~\ref{app:experimental_details}.
    
\item {\bf Code of ethics}
    \item[] Question: Does the research conducted in the paper conform, in every respect, with the NeurIPS Code of Ethics \url{https://neurips.cc/public/EthicsGuidelines}?
    \item[] Answer: \answerYes{} % Replace by \answerYes{}, \answerNo{}, or \answerNA{}.
    \item[] Justification: We have reviewed the NeurIPS Code of Ethics and followed appropriate ethical data-handling practices during benchmark construction. In particular, we defaced faces to protect privacy and excluded potentially harmful or inappropriate videos from the benchmark.

\item {\bf Broader impacts}
    \item[] Question: Does the paper discuss both potential positive societal impacts and negative societal impacts of the work performed?
    \item[] Answer: \answerYes{} % Replace by \answerYes{}, \answerNo{}, or \answerNA{}.
    \item[] Justification: We discuss both potential positive and negative societal impacts of our work in Appendix~\ref{app:societal_impact}.
    
\item {\bf Safeguards}
    \item[] Question: Does the paper describe safeguards that have been put in place for responsible release of data or models that have a high risk for misuse (e.g., pre-trained language models, image generators, or scraped datasets)?
    \item[] Answer: \answerYes{}
    \item[] Justification: We do not release any trained generative model or high-risk model. For the released benchmark data, we apply safeguards during construction, including face defacing for privacy and filtering out potentially harmful, inappropriate, or low-quality videos. We describe these data-handling procedures in Appendix~\ref{app:benchmark_detail}.

\item {\bf Licenses for existing assets}
    \item[] Question: Are the creators or original owners of assets (e.g., code, data, models), used in the paper, properly credited and are the license and terms of use explicitly mentioned and properly respected?
    \item[] Answer: \answerYes{}
    \item[] Justification: We credit existing assets and summarize the relevant license and terms-of-use considerations in Appendix~\ref{app:license_terms}.

\item {\bf New assets}
    \item[] Question: Are new assets introduced in the paper well documented and is the documentation provided alongside the assets?
    \item[] Answer: \answerYes{} % Replace by \answerYes{}, \answerNo{}, or \answerNA{}.
    \item[] Justification: We introduce new benchmark videos and code in this work. We provide the released data and code through the project URL, together with documentation and usage instructions.

\item {\bf Crowdsourcing and research with human subjects}
    \item[] Question: For crowdsourcing experiments and research with human subjects, does the paper include the full text of instructions given to participants and screenshots, if applicable, as well as details about compensation (if any)? 
    \item[] Answer: \answerNA{} % Replace by \answerYes{}, \answerNo{}, or \answerNA{}.
    \item[] Justification: This work does not involve crowdsourcing experiments or research with human subjects.

\item {\bf Institutional review board (IRB) approvals or equivalent for research with human subjects}
    \item[] Question: Does the paper describe potential risks incurred by study participants, whether such risks were disclosed to the subjects, and whether Institutional Review Board (IRB) approvals (or an equivalent approval/review based on the requirements of your country or institution) were obtained?
    \item[] Answer: \answerNA{} % Replace by \answerYes{}, \answerNo{}, or \answerNA{}.
    \item[] Justification: This work does not involve crowdsourcing experiments or research with human subjects.

\item {\bf Declaration of LLM usage}
    \item[] Question: Does the paper describe the usage of LLMs if it is an important, original, or non-standard component of the core methods in this research? Note that if the LLM is used only for writing, editing, or formatting purposes and does \emph{not} impact the core methodology, scientific rigor, or originality of the research, declaration is not required.
    %this research? 
    \item[] Answer: \answerNA{} % Replace by \answerYes{}, \answerNo{}, or \answerNA{}.
    \item[] Justification: LLMs are not used as an important, original, or non-standard component of the core methodology. Any use of LLMs is limited to writing, editing, or formatting assistance and does not affect the scientific content or experimental results.

\end{enumerate}
\fi

\end{document}